\documentclass[authoryear]{elsarticle}

\usepackage[brazilian,english]{babel}
\usepackage[utf8]{inputenc}
\usepackage[T1]{fontenc}

\usepackage{algorithm}
\usepackage{algpseudocode}
\usepackage{capt-of}
\usepackage{amsmath}
\usepackage{graphicx}
\usepackage{caption}
\usepackage{subcaption}
\usepackage{multirow}

\journal{Expert Systems with Applications}

\begin{document}

\begin{frontmatter}



\title{A Semi-Supervised Fuzzy GrowCut Algorithm to Segment and Classify Regions of Interest of Mammographic Images}


\author[a]{Filipe R. Cordeiro}
\ead{frc@cin.ufpe.br}

\author[b]{Wellington P. Santos\corref{cor1}}
\ead{wellington.santos@ufpe.br} 
\cortext[cor1]{Corresponding author.}

\author[a]{Abel G. Silva-Filho}
\ead{agsf@cin.ufpe.br}

\address[a]{Informatics Center, Federal University of Pernambuco, Brazil}
\address[b]{Department of Biomedical Engineering, Federal University of Pernambuco, Brazil}

\begin{abstract}
According to the World Health Organization, breast cancer is the most common form of cancer in women. It is the second leading cause of death among women round the world, becoming the most fatal form of cancer. Despite the existence of several imaging techniques useful to aid at the diagnosis of breast cancer, x-ray mammography is still the most used and effective imaging technology. Consequently, mammographic image segmentation is a fundamental task to support image analysis and diagnosis, taking into account shape analysis of mammary lesions and their borders. However, mammogram segmentation is a very hard process, once it is highly dependent on the types of mammary tissues. The GrowCut algorithm is a relatively new method to perform general image segmentation based on the selection of just a few points inside and outside the region of interest, reaching good results at difficult segmentation cases when these points are correctly selected. In this work we present a new semi-supervised segmentation algorithm based on the modification of the GrowCut algorithm to perform automatic mammographic image segmentation once a region of interest is selected by a specialist. In our proposal, we used fuzzy Gaussian membership functions to modify the evolution rule of the original GrowCut algorithm, in order to estimate the uncertainty of a pixel being object or background. The main impact of the proposed method is the significant reduction of expert effort in the initialization of seed points of GrowCut to perform accurate segmentation, once it removes the need of selection of background seeds. Furthermore, the proposed method is robust to wrong seed positioning and can be extended to other seed based techniques. These characteristics have impact on expert and intelligent systems, once it helps to develop a segmentation method with lower required specialist knowledge, being robust and as efficient as state of the art techniques. We also constructed an automatic point selection process based on the simulated annealing optimization method, avoiding the need of human intervention. The proposed approach was qualitatively compared with other state-of-the-art segmentation techniques, considering the shape of segmented regions. In order to validate our proposal, we built an image classifier using a classical multilayer perceptron. We used Zernike moments to extract segmented image features. This analysis employed 685 mammograms from IRMA breast cancer database, using fat and fibroid tissues. Results show that the proposed technique could achieve a classification rate of 91.28\% for fat tissues, evidencing the feasibility of our approach. 
\end{abstract}


\end{frontmatter}


\section{Introduction}
\label{sec:intro}

Breast cancer is the most common cancer in women worldwide: the World Health Organization (WHO) estimates the occurrence of 1.1 million new cases each year \citep{mathers2008global}. Survival rates for breast cancer can vary from 80\%, in high-income countries, to below 40\% in low-income nations \citep{coleman2008cancer}. The low survivability in some countries is related to the lack of screening programs which assist in the early detection of cancers. Early detection has an important impact on the successful treatment of cancer, once medical treatment becomes harder in late stages. One of the most effective methods for breast cancer analysis is digital mammography \citep{maitra2011identification}. However, mammography visual understanding and analysis can be a hard task even to a specialist, once such a procedure can be affected by image quality aspects, radiologist experience, and tumor shape.

A realistic estimative of the period that comprises the beginning of the tumor and its growth until it becomes palpable, reaching around 1 cm, is about 10 years \citep{allred1998prognostic}. During this period, breast imaging is essential for tumor monitoring. Correct evaluation of tumor size takes an important role at planning breast cancer treatments and avoiding mutilating surgeries, e.g. mastectomy \citep{litiere2012breast}. Nevertheless, imaging devices used by the BMH (Brazilian Ministry of Health) \citep{precoce2004controle} for the detection of breast cancer, which involve manual identification of the nodule size, are quite inefficient at the evaluation. These methods depend substantially on the professional examiners experience \citep{precoce2004controle} . Furthermore, image diagnosis is a complex task due to the large variability of clinical cases. Many cases seen in clinic practice do not fit classic images and descriptions precisely \citep{juhl2000interpretaccao}. For these reasons, mammography computer aided diagnosis (CAD) has been playing an import role to assist radiologists in improving the accuracy of their diagnosis. Consequently, traditional techniques in image processing have been applied in the medical field to make diagnosis less susceptible to errors through accurate identification of anatomic anomalies \citep{da2010algorithm}\citep{ye2010medical}.

The shape of the segmented tumor is a determinant factor in the mammogram diagnosis. It is related to the gravity of the tumor and the difference of a few centimeters in the maximum diameter can determine if it is necessary do a surgery or not. However, it can be very difficult to detect the contour of the tumor accurately depending on several factors, such as shape of the tumor, density, size, location and image quality. Some challenges in tumor segmentation include low contrast images, intensity levels which vary greatly across different regions, poor illumination and high noise levels, non-defined contours, and masses which are not always obviously detected \citep{raman2011review}.


The GrowCut algorithm is a relatively new method to perform general image segmentation based on the selection of just a few points inside and outside the region of interest, reaching good results at difficult segmentation cases when these points are adequately selected \ref{growcut}.

In this work we present a new semi-supervised segmentation algorithm based on the modification of the GrowCut algorithm to perform semi-automatic mammography image segmentation. In our proposal, we used fuzzy Gaussian membership functions to modify the evolution rule of the original GrowCut algorithm, in order to estimate the uncertainty of a pixel being object or background. Once point selection can be considered an important disadvantage of GrowCut, we also constructed an automatic point selection process based on the simulated annealing optimization method, avoiding the need of human intervention. The proposed approach was qualitatively compared with other state-of-the-art segmentation techniques, considering the shape of segmented regions. In order to validate our proposal, we built an image classifier using a classical multilayer perceptron. We used Zernike moments to extract segmented image features. This analysis employed 685 mammograms from IRMA breast cancer database, using fat and fibroid tissues.

This work has impact in the context of expert systems once it turns an expert system less dependent on the user knowledge, besides turning the process more robust to incorrect initialization. Moreover, the proposed method can be extended and applied to other expert systems, in other areas of application.

This work is organized as following: in section \ref{sec:related_work} we present the related work; in section \ref{sec:proposal} we present our segmentation proposal based on the modification of GrowCut algorithm using fuzzy Gaussian membership functions and the classical simulated annealing algorithm; in section \ref{sec:results} we present our experimental qualitative and quantitative results and perform some comments; finally, in section \ref{sec:conclusion} we present general conclusions and some perspectives of future works.

\section{Related Works} \label{sec:related_work}

Recent works have provided good accuracy in identifying the location of tumors \citep{liu2011new}\citep{mohamed2009mass}, however relatively little research has been done to verify the quality of segmentation. \cite{oliver2010review} makes a review of state of art and shows that related works are divided into edge-based segmentation, region-based segmentation and adaptive threshold.

In edge-based segmentation, it is difficult to determine the boundary of the tumor due to some ill-defined edges lesions. Region-based segmentation are more suitable for mass detection, since regions of tumor are usually brighter than their surrounding tissue, have an almost uniform density and a fuzzy boundary \citep{raman2011review}.

Recent studies for tumor segmentation have been successfully applied to region-based techniques for tumor segmentation.  \cite{lewis2012detection} uses Watershed to automatically segment tumor candidate regions, achieving an overall detection rate for mass tumors of 90\%. However, the metric of analysis that was used was based only on the location of the tumor and not on the quality of segmentation.

\cite{6708036} use an adaptive  threshold technique, achieving 100\% sensitivity, with an average of 1.87 false positives, when applied to 188 images. However, the value of sensitivity varies depending on the false positive rate, and each work uses a different rate.

Suspect regions usually are brighter than neighbor regions and with a uniform density \citep{5290137}. However, usually lesion regions do not have a well-defined contour. Due to this fact, seed-based techniques, i.e. techniques in which users label the initial seeds, show a better quality in the final segmentation. GrowCut technique has been applied to successful segment medical images, such as kidney \citep{6738236}, brain \citep{comparativemri} and vertebral body segmentation \citep{egger2013growcut}.  \cite{cordeiro2012segmentation} apply the classical GrowCut to segment masses in mammograms, obtaining good results in terms of quality of segmentation. \cite{zhengrandom} employ a random-walk based segmentation, which also uses seeds provided by the user, to achieve a good segmentation. However, they do not provide a quantitative analysis of the results. Despite seed based  techniques have shown suitable performance for mass segmentation, they require a high level of specialist knowledge  about the problem in order to select these seeds.

Unsupervised and Semi-Supervised techniques try to reduce the required specialist knowledge about the tumor region.  \cite{ghosh2011unsupervised} proposes an unsupervised GrowCut applied to medical images, but it is used for clustering and not for specific segmentation. Ramathi et. al use Active Contours \citep{rahmati2012mammography} to segment masses, achieving 86.85\% of accuracy using an overlap measure between segment images and ground truth. Chakraborty et al. apply Multilevel threshold \citep{chakraborty2012detection} combined with region growing to perform segmentation for well-defined edge contours, but both techniques show difficulties in defining spiculated contours or ill-defined edges.  \cite{hao2012automatic} attempt an automated seed generation combining isocontour maps with random walks and active contours, achieving high accuracy for the metric of area overlap measure. However, this metric alone does not reflect precisely the quality of segmentation.

\cite{al2015mammogram} et. al proposes an image visual enhancement and mass segmentation, obtaining tumor classification accuracy of 90.7\%. However, the segmentation step is mainly based on thresholding, which does not guarantee correct segmentation for ill-defined edges, even with image enhancement. \cite{dong2015efficient} proposes and automated for mass segmentation, using active contours to perform the segmentation. Nevertheless, it uses the information provided by the database to identify the location of the mass, which does not happens in practice. \cite {xie2016pcnn} use a Pulse Coupled Neural Network algorithm to obtain a scheme for correct initialization for level set evolution. However, the work does not explore the limitation of the algorithm to wrong initialization, once the level set segmentation depends on that. Although it improves the level set segmentation, the algorithm is still dependent of a good initialization.

As described previously, most of recent work in literature which are based on seeds selection are dependent on correct initialization in order to the algorithm perform accurately. But a correct seed positioning requires high user knowledge about the problem, to the most complex images. Although new methods with a high segmentation accuracy have been proposed, they are still high dependent on the user knowledge to obtain good results. The unsupervised methods proposed in literature have two main approaches: obtaining an automatic threshold value to perform the segmentation or generating the seeds automatically. The methods based on a threshold may have difficulties to perform segmentation in more complex images, with ill-defined edges. The techniques based on automatic seeding must guarantee that all the seeds are correctly positioned. The proposed method contributes and differs from state of art techniques by reducing the knowledge necessary to perform segmentation, using as case of study the GrowCut technique. The proposed techniques eliminate the need of selecting background seeds and makes the method more robust to wrong initialization. This has an impact of using unsupervised segmentation methods easier and more tolerant to different initializations. Furthermore, the proposed approach can be extended to other techniques and other kind of image.

As observed by Raman et al. \citep{raman2011review}, related works results differ significantly, and are often based on visual subjective opinion with very little quantitative endorsement. Furthermore, most studies describe an accuracy of the techniques based only on the localization of the tumor and not on its shape and contour, though these characteristics are very important for accurate diagnoses. Herein this work we propose a new approach based on automatic selection of seeds, making comparisons between our proposal and other state-of-the-art techniques, analyzing the quality of segmentation of each technique. 

\section{Methods} \label{sec:proposal}

In terms of the methodology, the segmentation can be thought of a process which consists of two tasks: the localization of the anatomy of interest and its delineation. The proposed methodology aims to provide assistance to the specialist to find an accurate delineation of the mass. Therefore, it assumed that a region of interest was previously selected by a specialist and provided to the proposed system to perform a high quality segmentation. Therefore, the objective of the proposed method is not to segment the mass from a full mammogram, but to help the professional to identify the correct measure of the mass. Once the region on interest (ROI) is used as input, the segmentation task is performed automatically. The method is called semi-supervised because of the need of selection of the region of interest by a specialist. But once the ROI is given as input to the proposed system, the segmentation is performed automatically.

\begin{figure}[!htb]
\centering
\includegraphics[width=\textwidth]{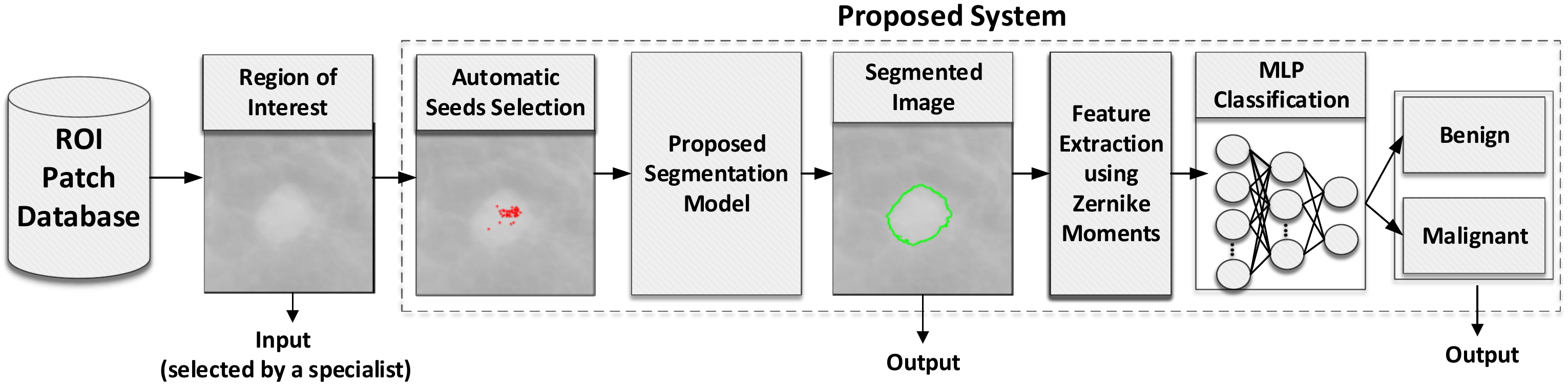}
\caption{Flowchart of the proposed method.}
\label{fig:flowchart}
\end{figure}

The flowchart of Figure \ref{fig:flowchart} illustrates the proposed method. 

The  methodology starts from an initial region of interest that corresponds to a previous selection made by a specialist or performed by a computational algorithm. In this work, ROIs are provided by the IRMA \citep{deserno2011towards} database, which contains patches of the suspicious image regions. After this, automatic selection of seeds is performed using Simulated Annealing \citep{dowsland2012simulated}, which models the localization of seed in an optimization problem, in which the objectives are to maximize the intensity of seed pixels and minimize the distance between them. After the seed pixels are obtained, they are used as inputs to the proposed segmentation model, in order to generate the segmented image. Therefore, once given a region of interest, the segmentation process is performed  automatically. As the IRMA database does not provide the ground truth of the images of interest, we decided to use a classifier to validate the segmentation through the identification of the segmented images based on their shape and edge characteristics. If the classifier is able to identify the type of tumor of segmented images accurately, we consider the segmentation suitable enough for the problem. Therefore, after the segmented images are obtained, the feature selection stage starts. In this stage, the proposed feature extractor calculates attributes related to shape and margin of the segmented regions using the Zernike Moments. Subsequently, a classifier is applied to identify the segmented images as benign or malignant tumors. After the classification step, we perform the analysis of results.

\subsection{Proposed Segmentation Model}

The proposed model is based on the GrowCut \citep{vezhnevets2005growcut} algorithm, a user interactive approach employed to perform image processing tasks, such as noise reduction and morphological and edge detection. GrowCut is a technique based on cellular automata \citep{hernandez1996cellular}, represented by grids of cells, where each cell can assume a finite number of states, which can vary according to the neighborhood rules. The neighborhood consists of a selection of neighbor pixels of a determined image, and can be defined by using Neumann and Moore neighborhood  models \citep{nayak2014survey}, for example. All the cells update their states according to the same update rule, based on the values of neighbor cells. Each time a rule is applied to a grid, a new iteration begins.

The GrowCut technique, as a user-interactive based approach, uses the concept of a seed pixel, in which the user initially labels a set of pixels in different classes of interest and, based on these seeds, the algorithm tries to label all the pixels of the image. In GrowCut, each cell has a strength value and, at each iteration, the neighbor  cells try to dominate this specific cell, changing its label. If a defender cell has a higher strength than its dominators, then it continues with the same label. Otherwise, the specific cell inherits the dominators' cell label. The process continues until the algorithm reaches convergence and all the cells stop changing their states. The pseudo-code of GrowCut is described in Algorithm \ref{growcut}.

\begin{algorithm}
\caption{GrowCut evolution rule}\label{growcut}
\begin{algorithmic}[1]
  \ForAll {$p \in P$}
  \State $l_{p}^{t+1} \gets l_p^{t}$ 
  \State $\Theta _{p}^{t+1} \gets \Theta_p^{t}$
 
  \ForAll {$q \in N(p)$}
      \If {$g(\left \| \vec{C_p} - \vec{C_q} \right \|_{2})\cdot \Theta_{q}^{t}>\Theta_{p}^{t}$} 
          \State  $l_{p}^{t+1} \gets l_{q}^{t}$
          \State  $\Theta_{p}^{t+1} \gets g(\left \| \vec{C_p} - \vec{C_q} \right \|_{2})\cdot \Theta_{q}^{t}$
      \EndIf
  \EndFor
 \EndFor
\end{algorithmic}
\end{algorithm}

According to the Algorithm \ref{growcut}, for each cell \emph{p} in a \emph{P} space of cells, previous states are copied, updating the label value of cell \emph{p} in iteration \emph{t+1}, represented as $l_{p}^{t+1}$, and the strength  value of cell p in iteration \emph{t+1}, as $\Theta_{p}^{t+1}$. Next, for each cell \emph{q} belonging to a neighbor of cell \emph{p}, represented as \emph{N(p)}, the update label condition is checked. In the condition of line 5, $\vec{C_p}$ and $\vec{C_q}$ are intensity vectors of the pixels \emph{p} and \emph{q} in the gray-scale space of colors, respectively, and $\Theta_{q}^{t}$ and $\Theta_{p}^{t}$ are values of strength of cells \emph{q} and \emph{p} in iteration \emph{t}. Function \emph{g}, in lines 5 and 7, is a decreasing monotonic function, represented by Equation \ref{eq:g}.

\begin{equation}
g(x) = \frac{1}{max\left \|\vec{C}  \right \|_2}
\label{eq:g}
\end{equation}

Finally, label and strength of cells are updated if the domination rule is satisfied, and the process repeats until the algorithm converges.

In GrowCut, as in the majority of seed-based techniques, the quality of segmentation depends directly on the positions of the initial seeds. Therefore, it depends on the user's knowledge to select appropriately seeds next to the edge of the object to be segmented. In the case in which some seeds are initially labeled incorrectly, the algorithm may perform an undesired and poor segmentation.

The proposed model aims to reduce the need for initial knowledge about the contour of the object, besides reducing the effort of selection of seeds. Moreover, the proposed model aims to be fault tolerant, allowing it to recover from incorrect seed selection.

In GrowCut all the initial seeds selected by the user have maximum strength value, assigning a high weight to the seeds with incorrect labels. Unlike GrowCut, the proposed model is based on the selection of seeds of only one class: the object of interest. The traditional GrowCut only  works with two classes, and if the background class is not close to the edges of tumor it does not provide a good segmentation, as described by Cordeiro et al. \citep{cordeiro2012segmentation}. In our approach, we discard the selection of a background class because, from the seeds of object class, we can estimate a frontier region separating object and background. However, instead of assigning all the labeled cells with maximum strength, all the cells are initialized with zero strength, except the cell corresponding to the center of mass of input seeds. Consequently, we assign maximum value to the cell of center of mass because  we assume that it has a higher chance of having a correct label. The initialization is performed using the following Equation \ref{eq:init}.

\begin{equation}
\forall p \in P,  l_{p} = 0, \Theta_{p} = 0, l_{cm} = l_{ob}, \Theta _{cm} = 1;
\label{eq:init}
\end{equation}

where \emph{p} is a cell in space \emph{P} of cells, and $l_{p}$ and $\Theta_{p}$ are the labels and strengths of cell \emph{p}, respectively. The label and strength of the cell which corresponds to the center of mass of the seeds are represented by $l_cm$ and $\Theta_cm$, respectively.

The proposed model makes a modification in the update rule of the cells of GrowCut, in a way that the attack of each cell is based in a region modeled by a Gaussian function. The strength of the model will be equal to 1 if the degree of membership of the specific cell to the background is higher than its complement, i.e. the degree of membership of the specific cell to the object of interest. Otherwise, the strength of the model assumes the strength of the current cell. The update algorithm of the proposed method is shown in Algorithm \ref{proposed}.

\begin{algorithm}
\caption{Proposed Algorithm evolution rule}\label{proposed}
\begin{algorithmic}[1]
  \ForAll {$p \in P$}
  \State $l_{p}^{t+1} \gets l_p^{t}$ 
  \State $\Theta _{p}^{t+1} \gets \Theta_p^{t}$
	\State Calculate $\Theta_{M,p}^t$
  \ForAll {$q \in N(p)$}
			\State Calculate $\Theta_{M,q}^t$
      \If {$g(\left \| \vec{C_p} - \vec{C_q} \right \|_{2})\cdot \Theta_{M,q}^t>\Theta_{M,p}^{t}$} 
          \State Calculate $l_{M,p,q}^t$
					\State  $l_{p}^{t+1} \gets l_{M,p,q}^{t}$
          \State  $\Theta_{p}^{t+1} \gets g(\left \| \vec{C_p} - \vec{C_q} \right \|_{2})\cdot \Theta_{M,q}^{t}$
      \EndIf
  \EndFor
 \EndFor
\end{algorithmic}
\end{algorithm}

In Algorithm \ref{proposed}, $\Theta_{M,p}^{t}$ and $\Theta_{M,q}^{t}$ are the strengths of the model for the cells \emph{p} and \emph{q}, respectively, being represented by Equations \ref{eq:mi} to \ref{eq:probobj}.

\begin{equation}
\Theta_{M,i}=\left\{
\begin{array}{ll}
1, & {\mu_\mathrm{Bkg} (i) > \mu_\mathrm{Obj} (i)} \\ 
\Theta_i, & {\mu_\mathrm{Bkg} (i) \leq \mu_\mathrm{Obj} (i)}\\
\end{array}
\right.,
\label{eq:mi}
\end{equation}

\begin{equation}
\mu_\mathrm{Bkg} (i) = 1 - \mu_\mathrm{Obj} (i),
\label{eq:probbg}
\end{equation}

\begin{equation}
\mu_\mathrm{Obj} (i) = \exp\left(-\frac{(x_i-x_m)^{2}}{2 \alpha_x  s_x^{2} }\right) \exp\left(-\frac{(y_i-y_m)^{2}}{2 \alpha_y s_y^{2} }\right),
\label{eq:probobj}
\end{equation}
where $\mu_\mathrm{Bkg} (i)$ is the the fuzzy membership degree associated to the uncertainty of the $i$-th cell belongs to the image background, whilst $\mu_\mathrm{Obj} (i)$ is the the fuzzy membership degree associated to the uncertainty of the $i$-th belongs to the object of interest. These fuzzy membership functions are Gaussian functions whose variables {$x_i$} and {$y_i$} correspond to the coordinates of the $i$-th cell in the grid, whereas {$x_m$} and {$y_m$} are the coordinates of the center of mass for the initially selected seeds; $s_x$ and $s_y$ are the standard deviation of initial points, whilst $\alpha_x$ and $\alpha_y$ are the weights of tuning of the Gaussian function, empirically determined according to the problem of interest.

The label of each $q$-th cell, $l_{M,p,q}$, is updated according to the following expression of Equation \ref{eq:lmq}
\begin{equation}
l_{M,p,q}=\left\{
\begin{array}{ll}
l_{p}, & {\mu_\mathrm{Bkg} (q) > \mu_\mathrm{Obj} (q)} \\ 
l_{q}, & {\mu_\mathrm{Bkg} (q) \leq \mu_\mathrm{Obj} (q)}\\
\end{array}\right..
\label{eq:lmq}
\end{equation}

Table \ref{comparison} makes a comparison between the GrowCut algorithm and the proposed model.

\begin{table}[htb]
\centering
\caption{Comparison between GrowCut and the Proposed Algorithm.}
\label{comparison}
\footnotesize
\begin{tabular}{p{0.2\textwidth}|p{0.35\textwidth}p{0.35\textwidth}}
	\hline
	\textbf{Characteristic} & \textbf{GrowCut} & \textbf{Proposed Model}\\ 
	\hline
	Selection of Seeds & Selection of seeds of object class and background class. & Selection of seed only of object class.\\
	\hline
	Initialization & All the seeds have strength value equal to 1. & Only the cell corresponding to the center of mass of points has strength  value equal to 1.\\
	\hline
	Segmentation & Based on knowledge of seeds localization provided by the user. & Based on knowledge of seeds; localization and in the Gaussian model that separates the region of foreground and background region.\\
	\hline
	Fault Tolerance to seeds localization & Low & High\\   
	\hline
\end{tabular}
\end{table}

The initial impact of the proposed approach is the reduction of the effort to select the initial seeds, in which it is necessary to use only the seeds of the object of interest. The background region is obtained through the Gaussian model, regulating the strength of each cell in the update labeling process. The Gaussian model allows the process to be tolerant to incorrect selection of initial seeds, once it is based on the center of mass of the seeds. Consequently, the algorithm becomes less dependent on the user specialist knowledge, being more appropriate for the process of semi-supervised seed selection.

\subsection{Automatic Selection of Seeds}

The selection of seeds consists of identifying initial pixels located in regions of tumor and non-tumor. In many seed-based techniques, such as Random Walks \citep{grady2006random} and Graph Cut \citep{vicente2008graph}, seeds are selected manually by a specialist. In this work, the technique Simulated Annealing \citep{dowsland2012simulated} is used to automatically find the seeds in the region of interest. As usually, mass regions have higher intensity pixel values. Therefore, the problem of finding a set of seeds was converted into an optimization problem, where the algorithm optimizes the set of seeds by the intensity values, aiming to get the seed inside the mass areas. As the Simulated Annealing is a validated optimization algorithm, it is used to find a set of seeds, trying to minimize the fitness function described by equation \ref{eq:xdef}:

\begin{equation} 
fitness = \alpha \sum_{j=1}^{n-1}d_{jn} - \beta \sum_{j=1}^{n} I_j,
\label{eq:xdef}
\end{equation}

where $d_{jn}$ is the Euclidian distance between seed $j$ and seed $n$, and $I_j$ is the intensity value of seed $j$. Hence, the fitness function evaluates the intensity levels of the set of seeds and the distance between them. As it is also important that the seeds are spread throughout the region of interest, the distance between points is evaluated in the fitness function. Parameters $\alpha$ and $\beta$ are used to adjust the impact of distance and intensity of the seeds, respectively. The higher the value, the higher the influence of the distance of intensity in the fitness function. However, we recommend the values to be between 1 and 2. For the present application, we empirically defined the following standard values: $\alpha=1$ and $\beta=1.5$. An important aspect of the proposed algorithm is that it is not necessary to select non-tumor seeds, once our algorithm can adjust its fuzzy Gaussian frontier based only on the seeds of the tumor region.

\begin{figure}[!htb]
	\centering
 	\begin{subfigure}[b]{0.13\textwidth}
		\caption{}
		\includegraphics[width=\textwidth]{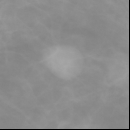}
		\includegraphics[width=\textwidth]{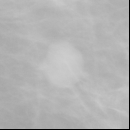}
		\includegraphics[width=\textwidth]{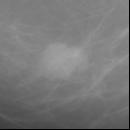}
		\includegraphics[width=\textwidth]{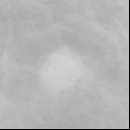}
		\includegraphics[width=\textwidth]{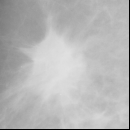}
		\includegraphics[width=\textwidth]{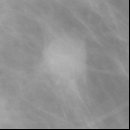}
	 	\includegraphics[width=\textwidth]{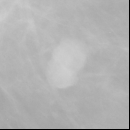}
	\end{subfigure}%
	\
 	\begin{subfigure}[b]{0.13\textwidth}
		\caption{}
		\includegraphics[width=\textwidth]{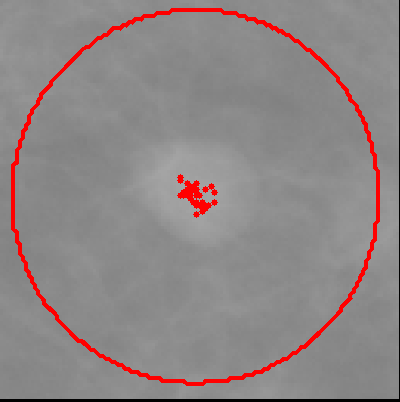}
		\includegraphics[width=\textwidth]{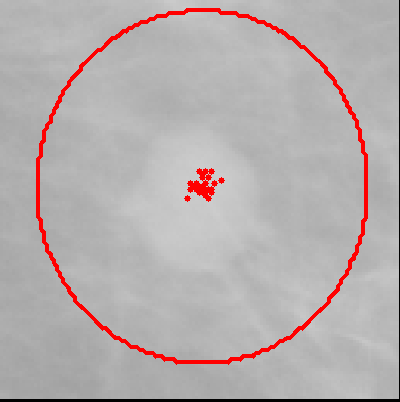}
		\includegraphics[width=\textwidth]{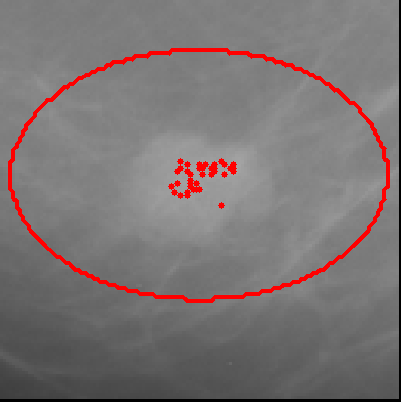}
		\includegraphics[width=\textwidth]{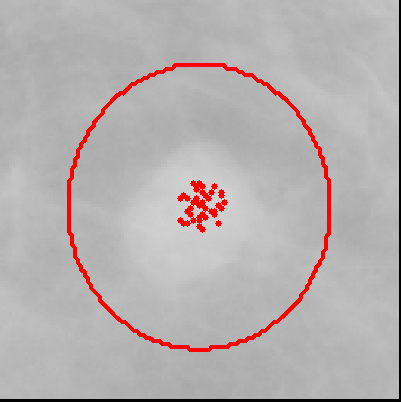}
		\includegraphics[width=\textwidth]{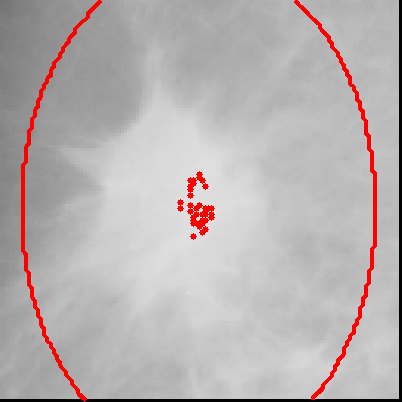}
		\includegraphics[width=\textwidth]{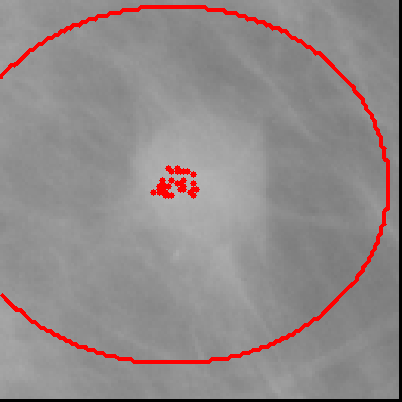}
	 	\includegraphics[width=\textwidth]{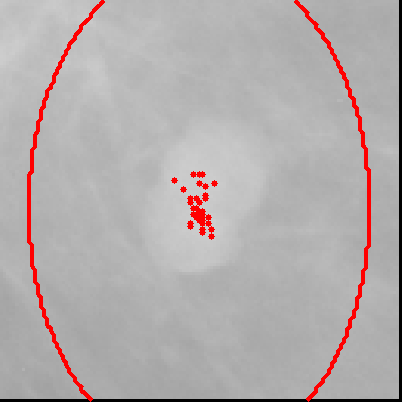}
	\end{subfigure}%
        \
	\begin{subfigure}[b]{0.13\textwidth}
		\caption{}
		\includegraphics[width=\textwidth]{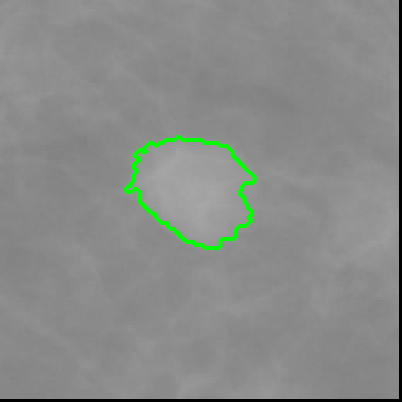}
		\includegraphics[width=\textwidth]{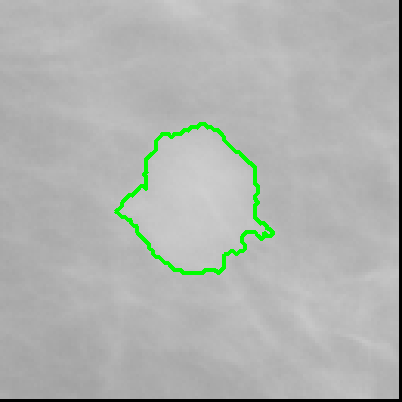}
		\includegraphics[width=\textwidth]{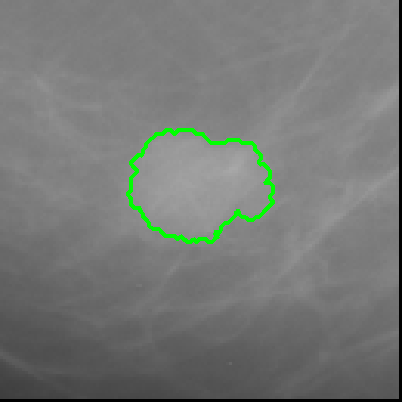}
		\includegraphics[width=\textwidth]{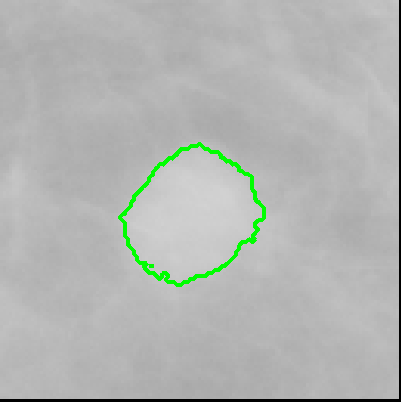}
		\includegraphics[width=\textwidth]{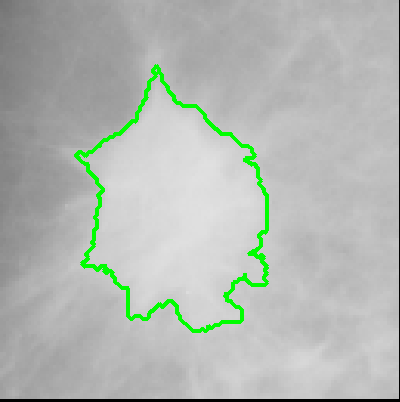}
		\includegraphics[width=\textwidth]{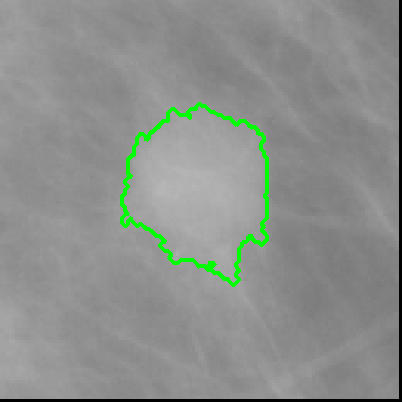}
	 	\includegraphics[width=\textwidth]{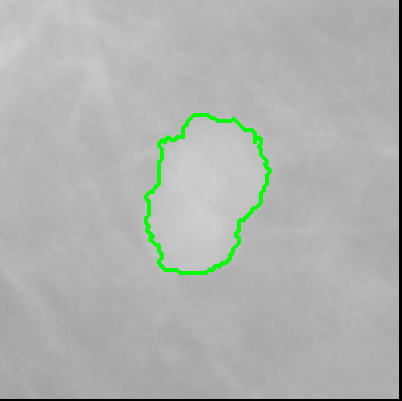}
	\end{subfigure}%
  \  
  \begin{subfigure}[b]{0.13\textwidth}
		\caption{}
		\includegraphics[width=\textwidth]{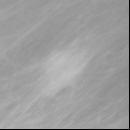}
		\includegraphics[width=\textwidth]{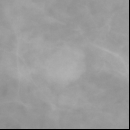}
		\includegraphics[width=\textwidth]{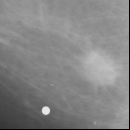}
		\includegraphics[width=\textwidth]{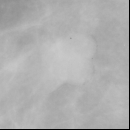}
		\includegraphics[width=\textwidth]{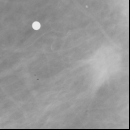}
		\includegraphics[width=\textwidth]{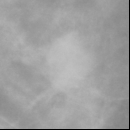}
	 	\includegraphics[width=\textwidth]{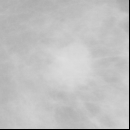}
	\end{subfigure}%
	\
 	\begin{subfigure}[b]{0.13\textwidth}
		\caption{}
   	\includegraphics[width=\textwidth]{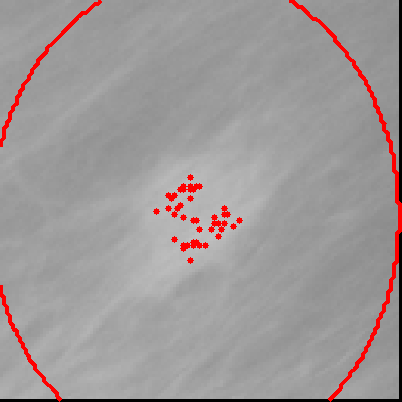}
		\includegraphics[width=\textwidth]{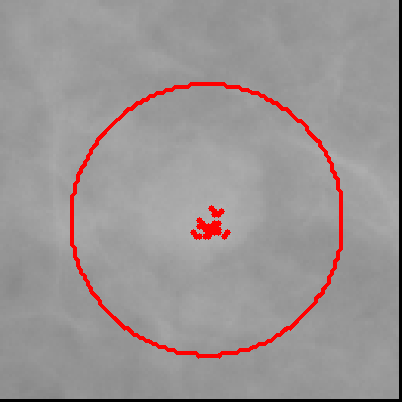}
		\includegraphics[width=\textwidth]{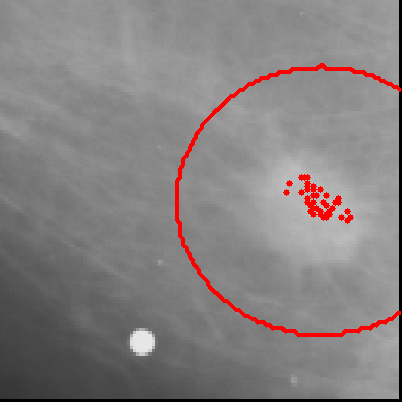}
		\includegraphics[width=\textwidth]{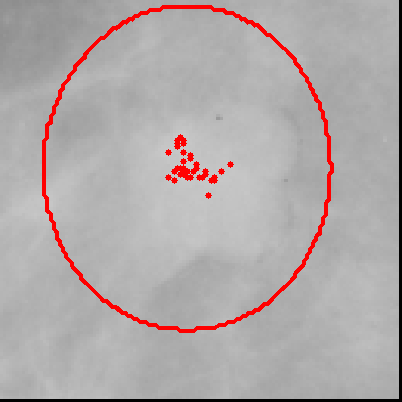}
		\includegraphics[width=\textwidth]{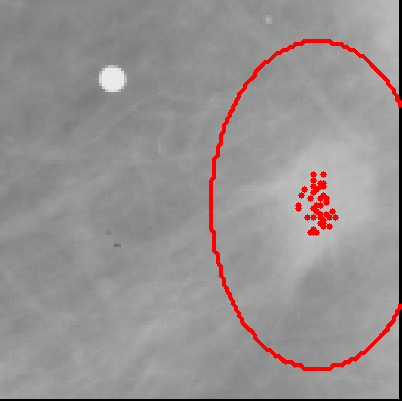}
		\includegraphics[width=\textwidth]{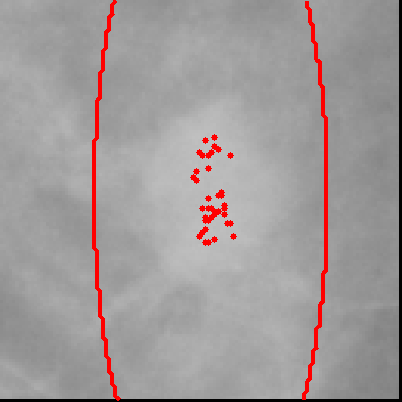}
	 	\includegraphics[width=\textwidth]{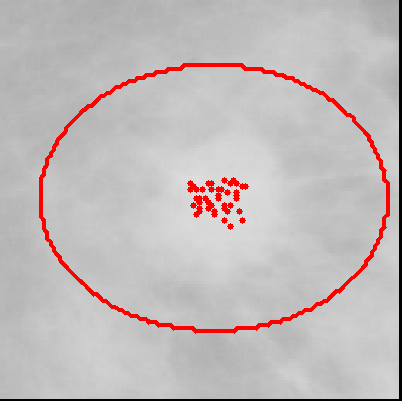}
	\end{subfigure}%
	\
	\begin{subfigure}[b]{0.13\textwidth}
		\caption{}
		\includegraphics[width=\textwidth]{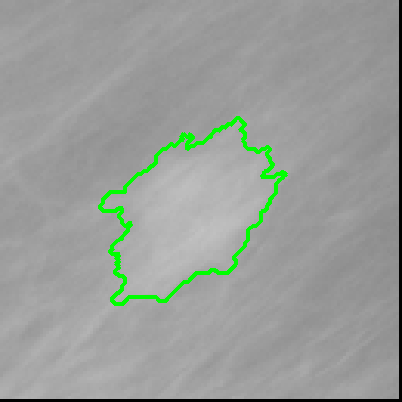}
		\includegraphics[width=\textwidth]{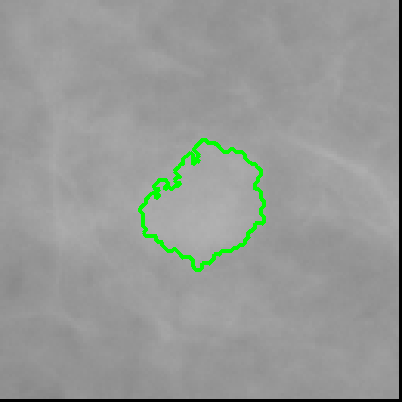}
		\includegraphics[width=\textwidth]{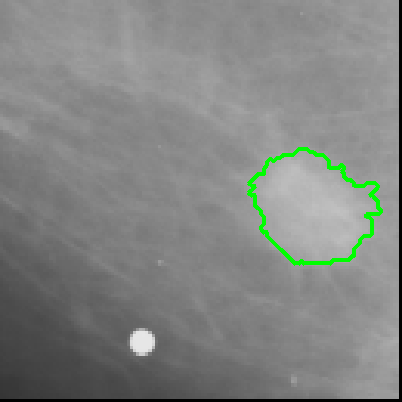}
		\includegraphics[width=\textwidth]{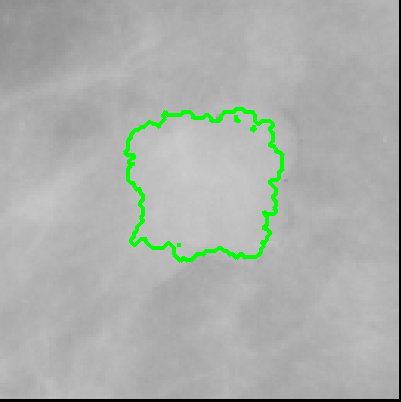}
		\includegraphics[width=\textwidth]{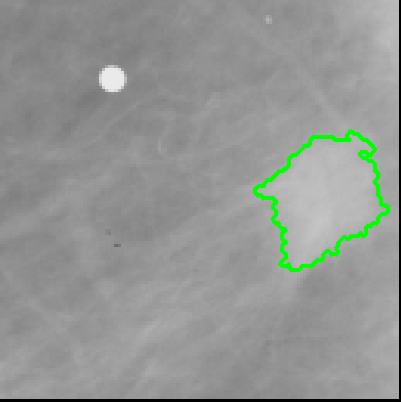}
		\includegraphics[width=\textwidth]{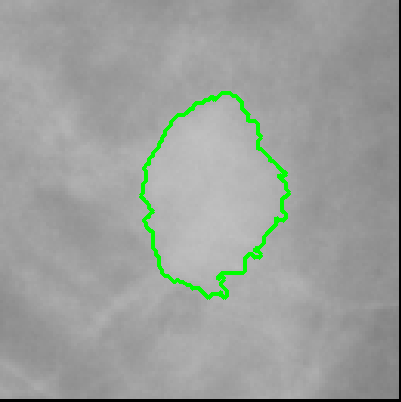}
	 	\includegraphics[width=\textwidth]{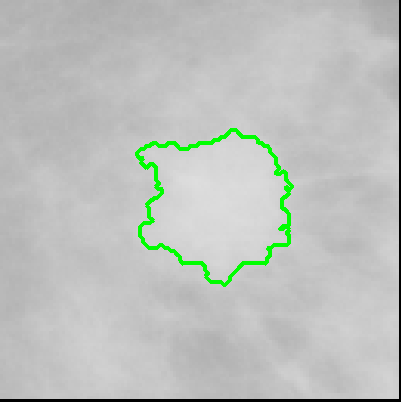}
	\end{subfigure}%
	\caption{Segmentation process of the proposed approach for images of IRMA database.  (a) and (d) Original Images; (b) and (e) Automatic generated seeds and Fuzzy-Gaussian region; (c) and (f) Final segmentations.}
	\label{fig:process}
\end{figure}

Figure \ref{fig:process} illustrates the steps of the segmentation process for some images of the database. Columns \emph{a} and \emph{d} of Figure \ref{fig:process} represents the initial region of interest selected from the IRMA database. Columns \emph{b} and \emph{e} shows the seed points obtained from automatic seeds selection, represented by the red points, and the fuzzy Gaussian region, represented by red ellipses. Regions inside ellipse have a higher probability of finding pixels of tumor mass. The size of the Gaussian region is based on the location and distribution of the seed points. The advantage of the proposed technique is that it requires only seeds of the tumor region, different from most of techniques. Finally, columns \emph{c} and \emph{f} shows the final segmentation of the proposed approach, represented by green contours. 

\section{Experiments}

\subsection{Experimental Environment}

Our proposal was evaluated using the IRMA \citep{deserno2011towards} \citep{de2010mammosys} \citep{deserno2012computer} database, which was developed from a project performed by Aachen University (RWTH Aachen). The database is composed by regions of interest of mammograms, which were classified by radiologists and resized to 128 $\times$ 128 pixels. The database is composed by 2.796 mammograms images of four repositories: 150 images from Mini-MIAS database \citep{suckling1994mammographic}, 2.576 images from DDSM \citep{heath2000digital}, 1 from LLN database, and 69 from RWTH database. The images from IRMA have four types of tissue density, which are classified in four types, according to the classification of BI-RADS \citep{d1998breast}: fat tissue (Type I), fibroid tissue (Type II), heterogeneous dense tissue (Type III) and extremely dense tissue (Type IV). In this work, we analyzed images of fat transparent and fibroid glands systems, for masses classified as circumscribed, spiculated, and other mass, according to the database description. For experimental evaluations, we used 685 mammography patches, which corresponds to all images of fat and fibroid tissues which lesions of type circumscribed, spiculated and other mass.

\subsection{Feature Extraction}
Feature extraction used in this work is based on the calculation of Zernike Moments \citep{tahmasbi2011classification}. The Zernike Moments are image descriptors of shape and margin and invariant to rotation, non-redundant, and robust to noise and shape \citep{wang2009mode}\citep{hwang2006novel}, and they had already been used successfully to identify masses by Tahmasbi et al. \citep{tahmasbi2011classification}. 

The Zernike Moments are defined as projections of the intensity function of an image, represented by $f:S\rightarrow W$, over the orthogonal basis functions, which are the Zernike polynomials. The calculation of Zernike Moments to a digital image $f$ is represented by Equation \ref{eq:z2}.

\begin{equation}
Z_{n,m}=\frac{n+1}{\pi (N-1)}\sum_{u\in S}f(u) V_{n,m}(\rho ,\theta ),
\label{eq:z2}
\end{equation}

where $\rho=\frac{\sqrt{x^2+y^2}}{N}$ and $\theta=tan^-1 (y/x)$. The variable $n$ is a natural number denominated moment order and $m$ is a positive or negative integer, named repetition, which satisfies the restriction $n-|m|=pair$, and $|m|\leq n$. The variable $V_{n,m}$ is the Zernike polynomials family, defined by the Equation \ref{eq:z3} and Equation \ref{eq:z1}.

\begin{equation}
V_{n,m}(\rho, \theta)=R_{n,m}(\rho)^{-jm\theta},
\label{eq:z3}
\end{equation}

\begin{equation}
R_{n,m}=\sum_{s=0}^{\frac{n-|m|}{2}}(-1)^s \frac{(n-s)!}{s!(\frac{n+|m|}{2}-s)!(\frac{n-|m|}{2}-s)!}\rho^{n-2s}.
\label{eq:z1}
\end{equation}

To calculate the Zernike Moments of an image, its center is considered as a center of an unitary disk. The Zernike Moments are divided in 64 descriptors, which are divided in two groups of 32 elements, defined as low order and high order moments.

\subsection{Classification}
In order to perform the classification of the suspicious regions of interest, we used a classical Multilayer Perceptron (MLP) \citep{jain1996artificial}, which is an extensively validated neural network based classifier. The inputs of the MLP are the Zernike moments extracted from the segmented images. We employed the following architecture: 64 inputs, two neurons in the output layer (benign and malignant finding classes), and two hidden layers with 30 neurons each one.

Training and test stages were performed using k-fold cross-validation, with 10 folds. The classifier was used to indirectly evaluate the quality of segmentation through the features of shape and margin extracted using Zernike moments. 

\subsection{Evaluation}

We evaluate the quality of segmentation of the implemented techniques by analyzing if the contour of the segmentation is well-defined enough to makes possible the correct identification of the type of tumor using the MLP classifier. We chose this evaluation because the IRMA database does not provide the segmentation ground truth. Furthermore, it would be unfeasible to a specialist manually segment all the images. Additionally, if the contour of the segmentation is suitable enough to turns possible the classifier identify the type of tumor, we can consider the segmentation has a good quality and is useful to be employed in clinical practice.

Our proposal was compared to six state-of-the-art works: BEMD \citep{jai2013mass}, BMCS \citep{berber2013breast}, LBI \citep{sharma2013roi}, MCW \citep{lewis2012detection}, Topographic Approach \citep{5290137} and Wavelet Analysis \citep{pereira2014segmentation}. Each technique was implemented using the parameters provided by each article. Although some works were used for different databases and considering the full mammogram, the tuning was based on the parameters suggest in each work.

A comparison with the classical GrowCut was not feasible due to necessity of selection of seed points of 685 images, which the database does not provide the ground truth. Furthermore, it is more suitable the comparison between semi-supervised techniques, as evaluated in this work.

\section{Results} \label{sec:results}

\begin{figure}[!htb]
	\centering
 	\begin{subfigure}[b]{0.110\textwidth}
		\caption{}
		\includegraphics[width=\textwidth]{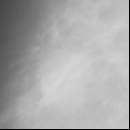}
		\includegraphics[width=\textwidth]{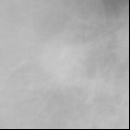}
		\includegraphics[width=\textwidth]{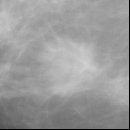}
		\includegraphics[width=\textwidth]{figures/original/1003071_cut_803774.png}
		\includegraphics[width=\textwidth]{figures/original/1003566_cut_805320.png}
	 	\includegraphics[width=\textwidth]{figures/original/1002694_cut_804626.png}
		\includegraphics[width=\textwidth]{figures/original/1002743_cut_805887.png}
		\includegraphics[width=\textwidth]{figures/original/1002794_cut_804183.png}
		
	\end{subfigure}%
	\
  \begin{subfigure}[b]{0.110\textwidth}
		\caption{}
		\includegraphics[width=\textwidth]{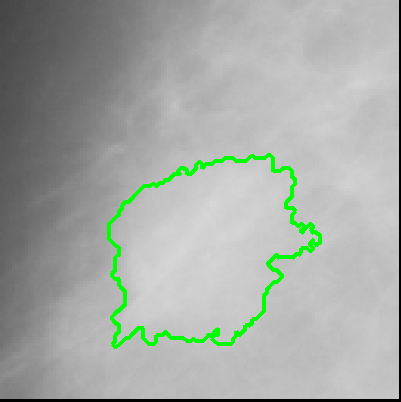}
		\includegraphics[width=\textwidth]{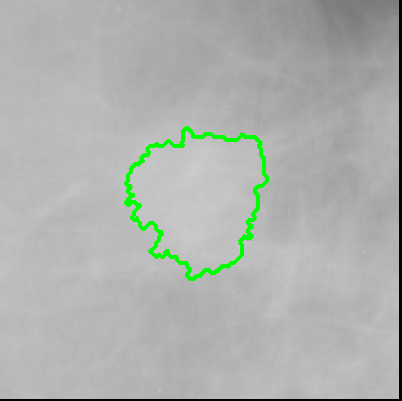}
		\includegraphics[width=\textwidth]{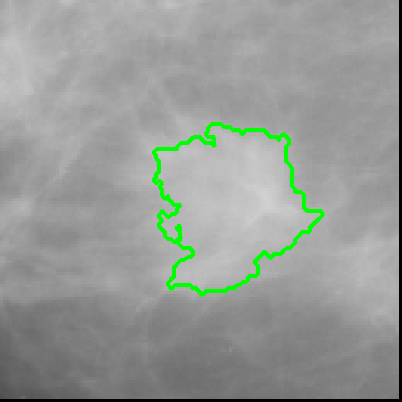}
		\includegraphics[width=\textwidth]{figures/gauss/tp1_1003071_seg.png}
		\includegraphics[width=\textwidth]{figures/gauss/tp1_1003566_seg.png}
	 	\includegraphics[width=\textwidth]{figures/gauss/tp2_1002694_seg.png}
		\includegraphics[width=\textwidth]{figures/gauss/tp2_1002743_seg.png}
		\includegraphics[width=\textwidth]{figures/gauss/tp2_1002794_seg.png}
		
	\end{subfigure}%
	\
	\begin{subfigure}[b]{0.110\textwidth}
		\caption{}
		\includegraphics[width=\textwidth]{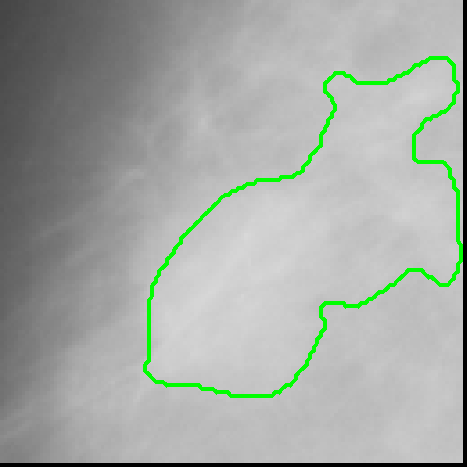}
		\includegraphics[width=\textwidth]{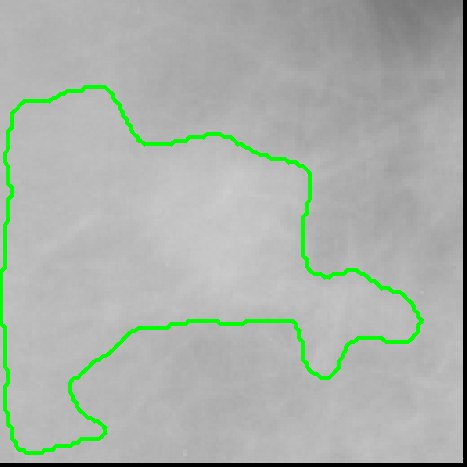}
		\includegraphics[width=\textwidth]{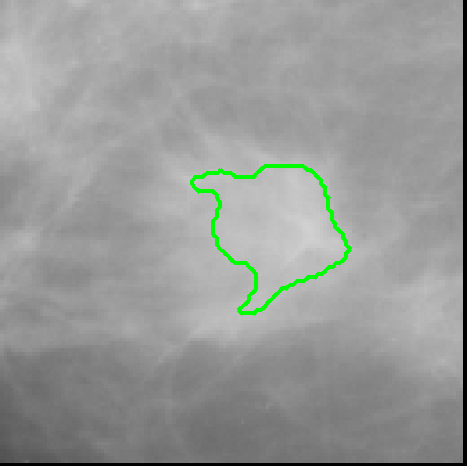}
		\includegraphics[width=\textwidth]{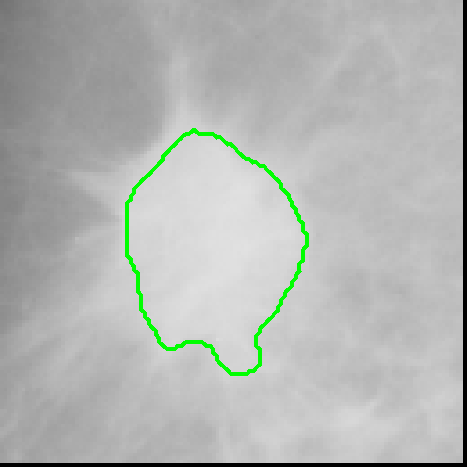}
		\includegraphics[width=\textwidth]{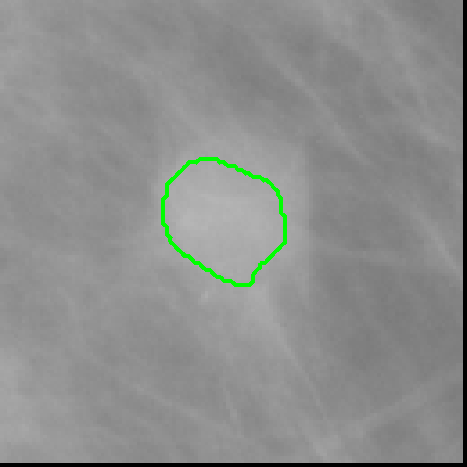}
	 	\includegraphics[width=\textwidth]{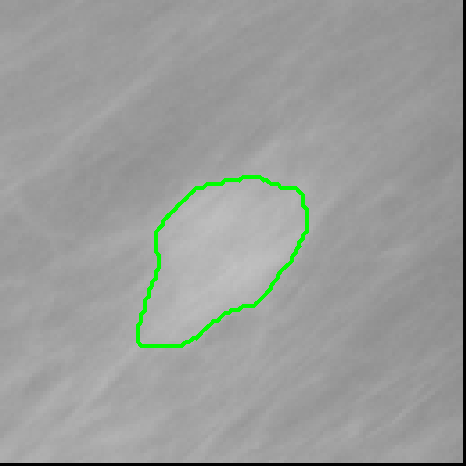}
		\includegraphics[width=\textwidth]{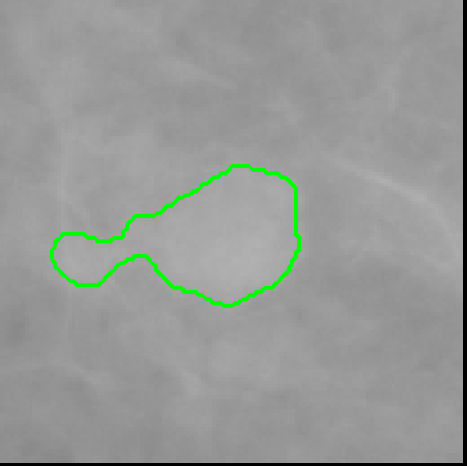}
		\includegraphics[width=\textwidth]{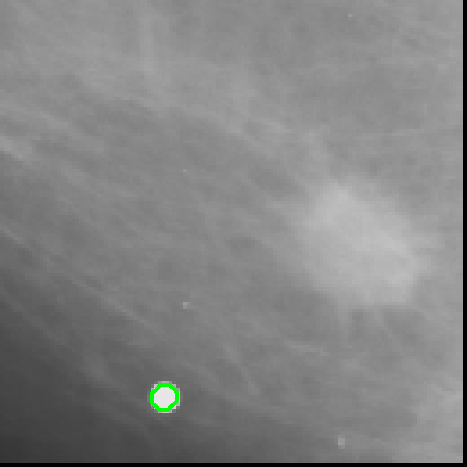}
		
	\end{subfigure}
	\
	\begin{subfigure}[b]{0.110\textwidth}
		\caption{}
		\includegraphics[width=\textwidth]{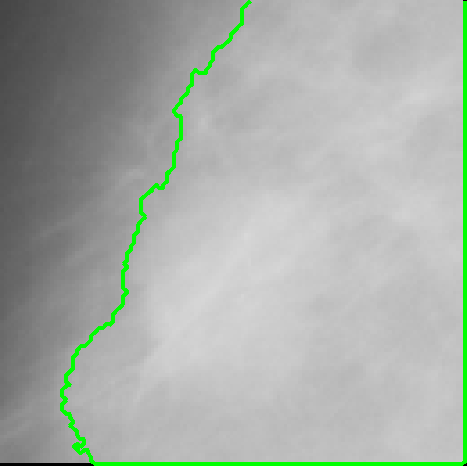}
		\includegraphics[width=\textwidth]{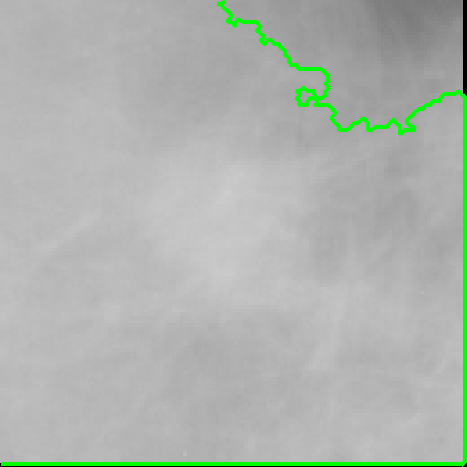}
		\includegraphics[width=\textwidth]{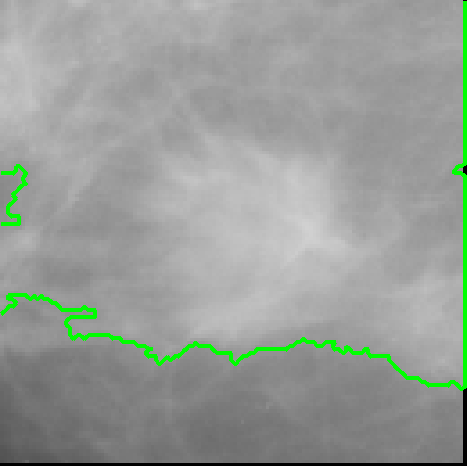}
		\includegraphics[width=\textwidth]{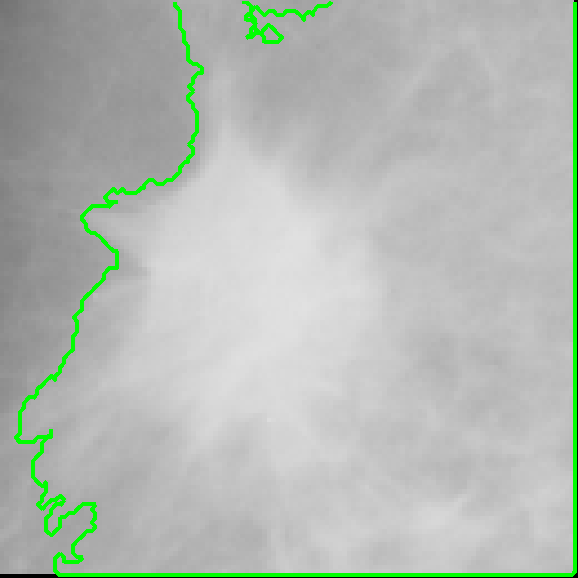}
		\includegraphics[width=\textwidth]{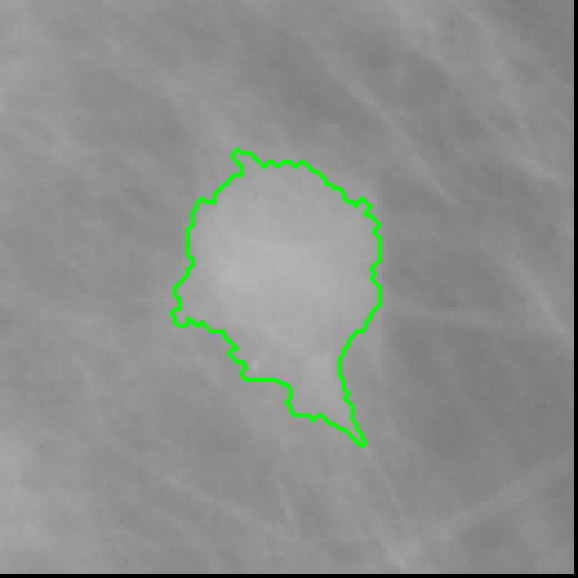}
		\includegraphics[width=\textwidth]{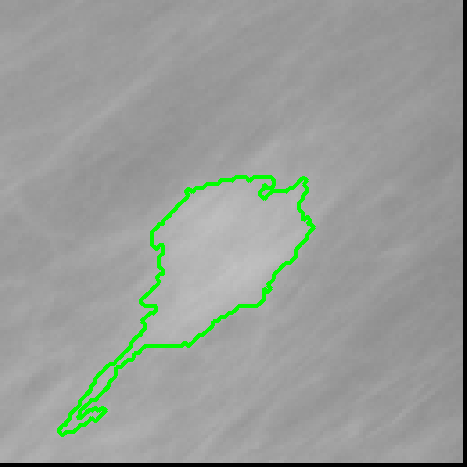}
		\includegraphics[width=\textwidth]{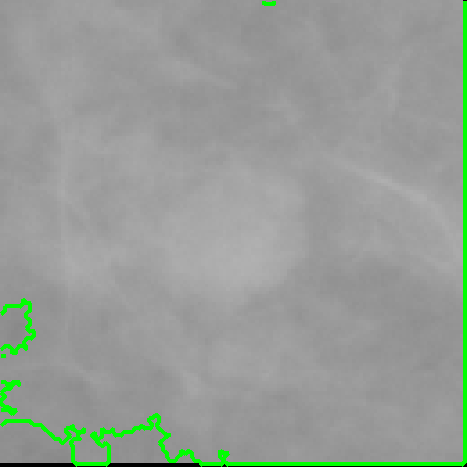}
		\includegraphics[width=\textwidth]{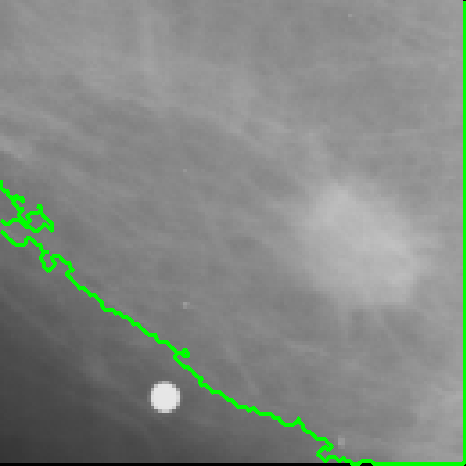}
	\end{subfigure}
	\
	\begin{subfigure}[b]{0.110\textwidth}
		\caption{}
		\includegraphics[width=\textwidth]{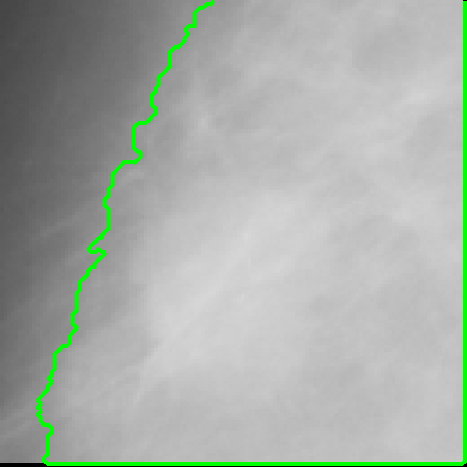}
		\includegraphics[width=\textwidth]{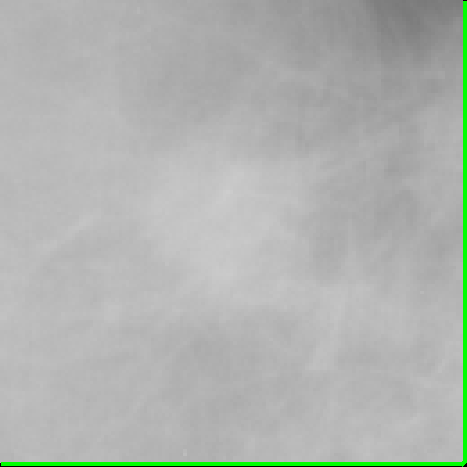}
		\includegraphics[width=\textwidth]{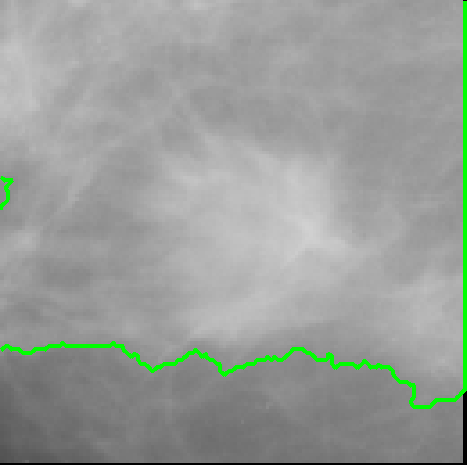}
		\includegraphics[width=\textwidth]{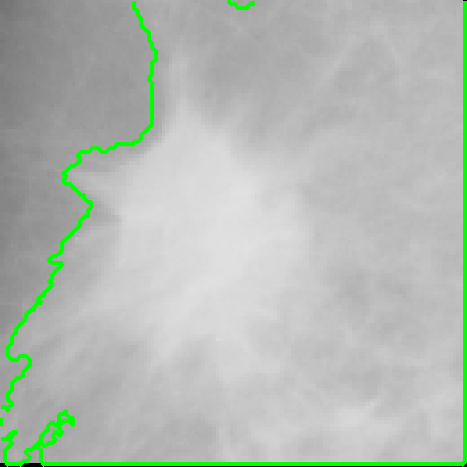}
		\includegraphics[width=\textwidth]{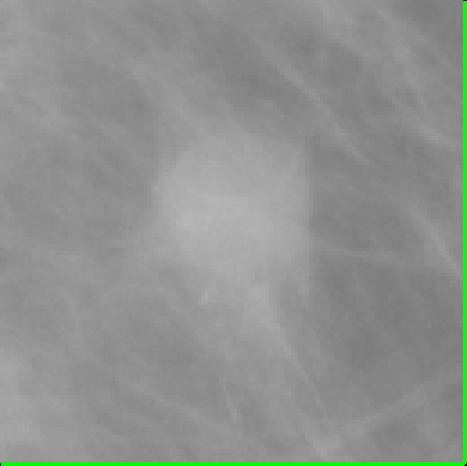}
		\includegraphics[width=\textwidth]{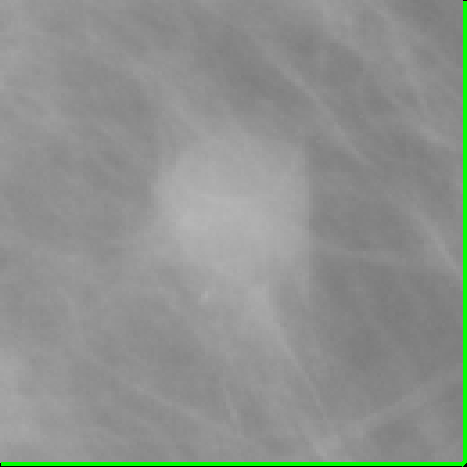}
		\includegraphics[width=\textwidth]{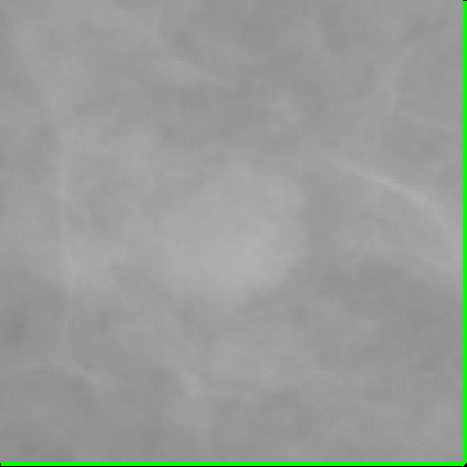}
		\includegraphics[width=\textwidth]{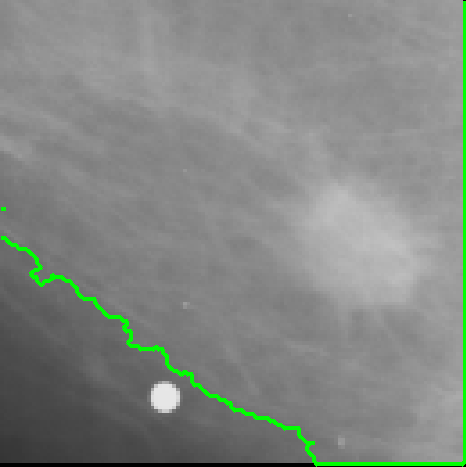}
	\end{subfigure}
	\
	\begin{subfigure}[b]{0.110\textwidth}
		\caption{}
		\includegraphics[width=\textwidth]{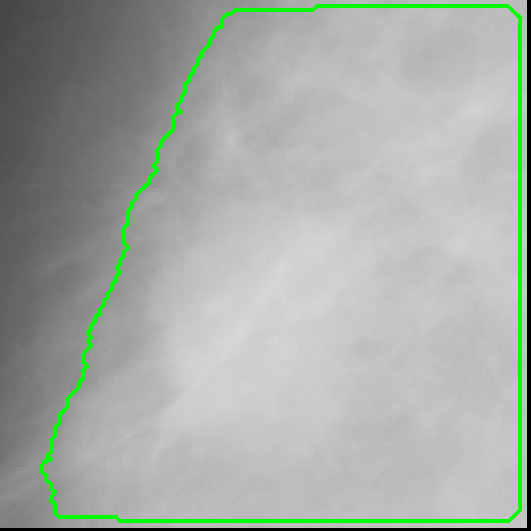}
		\includegraphics[width=\textwidth]{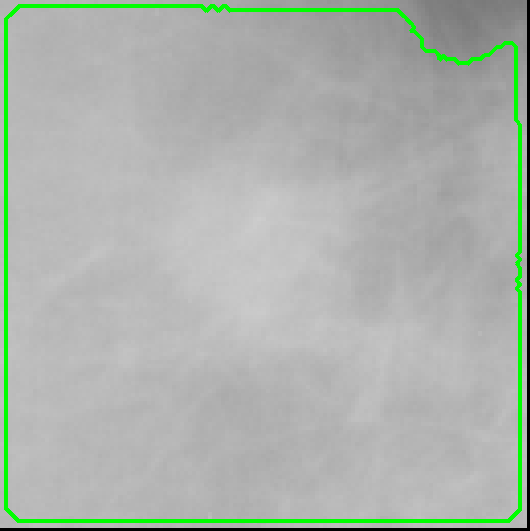}
		\includegraphics[width=\textwidth]{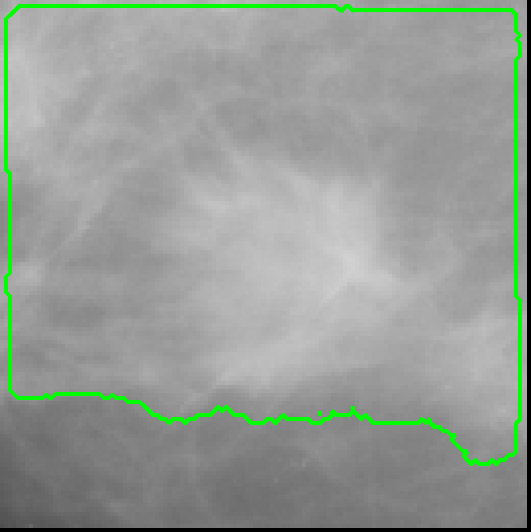}
		\includegraphics[width=\textwidth]{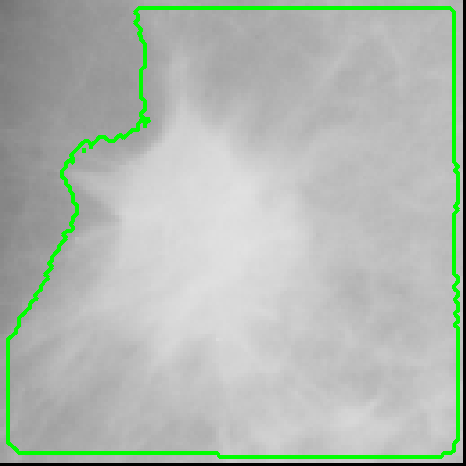}
		\includegraphics[width=\textwidth]{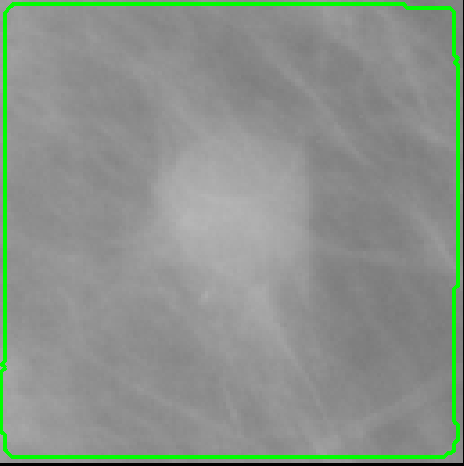}
		\includegraphics[width=\textwidth]{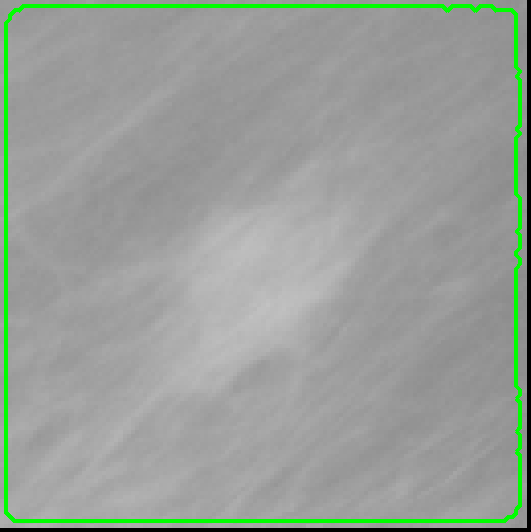}
		\includegraphics[width=\textwidth]{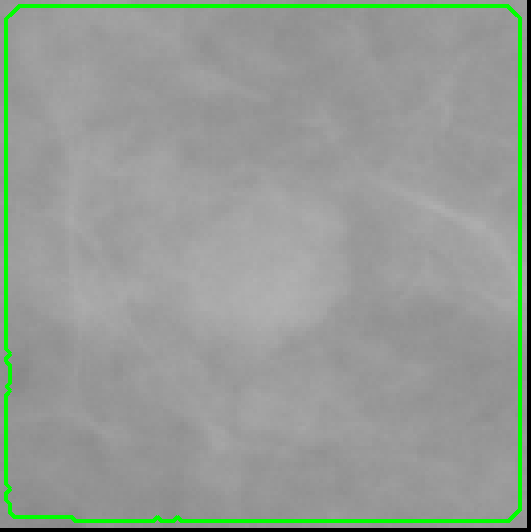}
		\includegraphics[width=\textwidth]{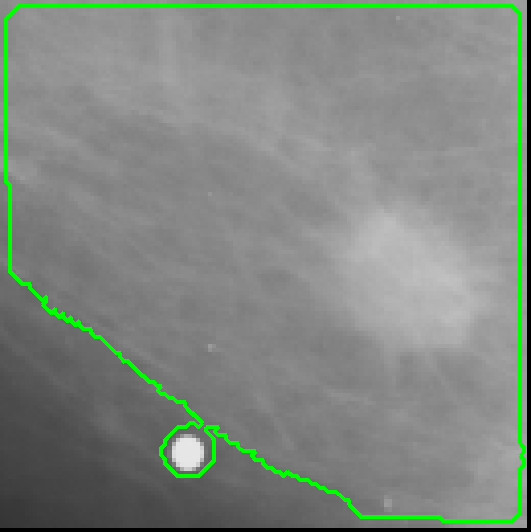}
	\end{subfigure}
	\
  \begin{subfigure}[b]{0.110\textwidth}
		\caption{}
    \includegraphics[width=\textwidth]{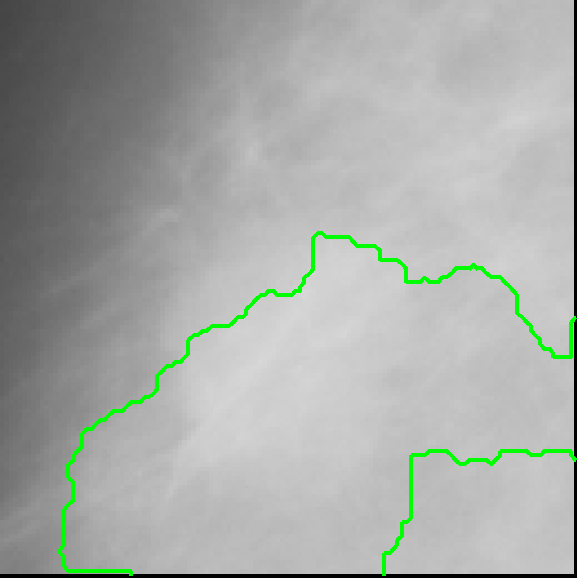}
		\includegraphics[width=\textwidth]{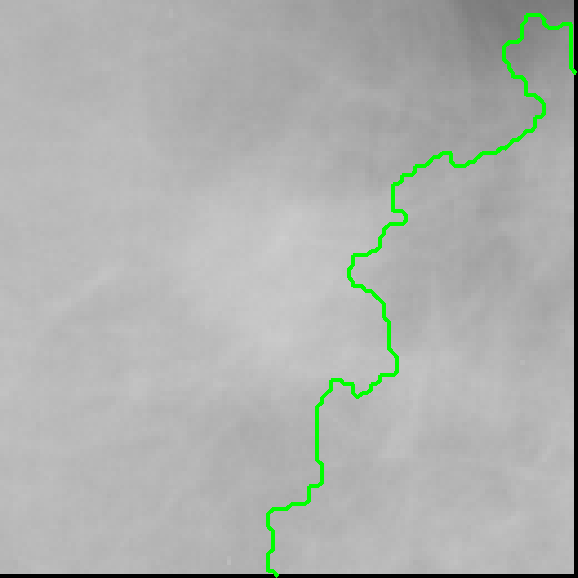}
		\includegraphics[width=\textwidth]{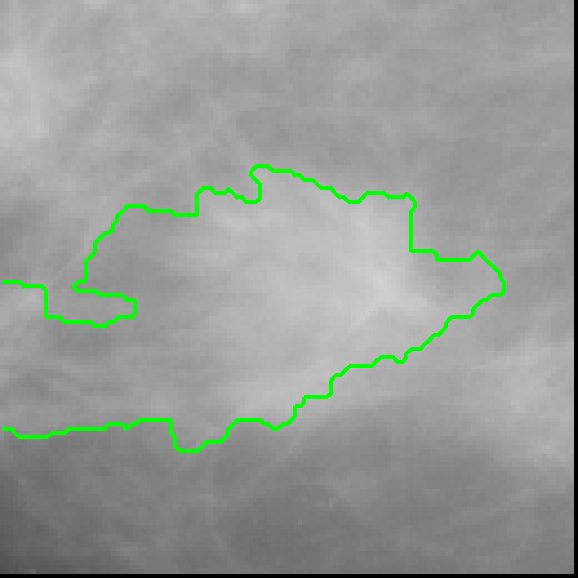}
		\includegraphics[width=\textwidth]{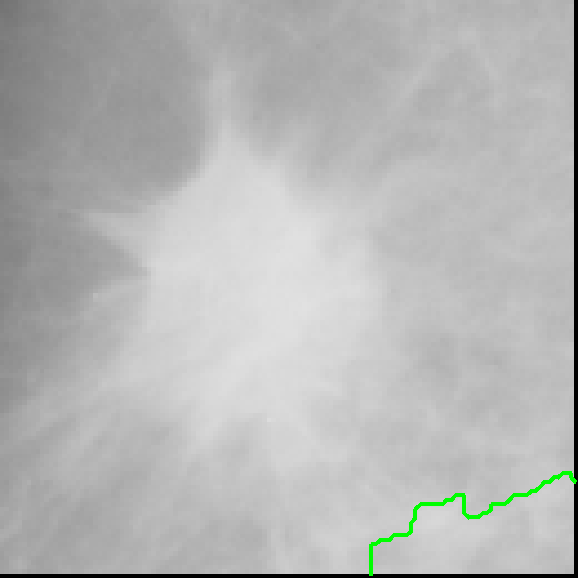}
		\includegraphics[width=\textwidth]{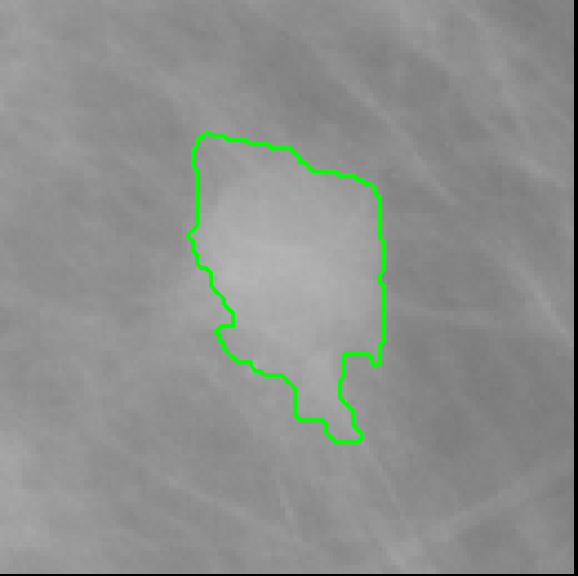}
		\includegraphics[width=\textwidth]{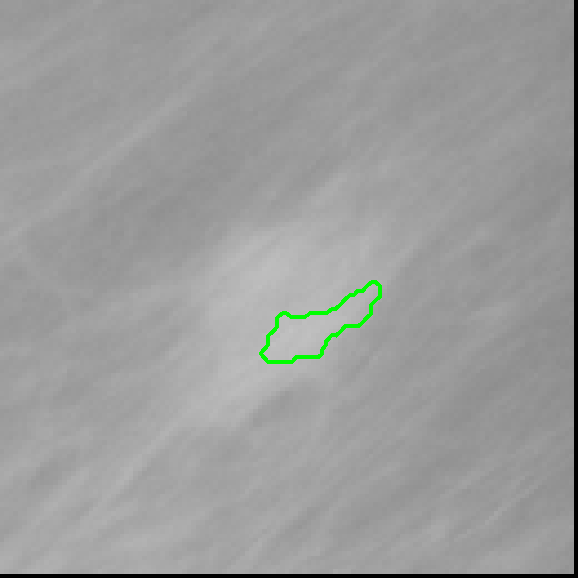}
    \includegraphics[width=\textwidth]{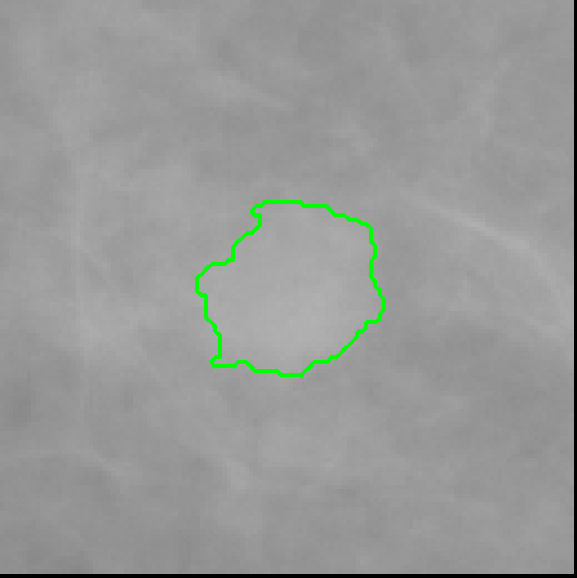}
		\includegraphics[width=\textwidth]{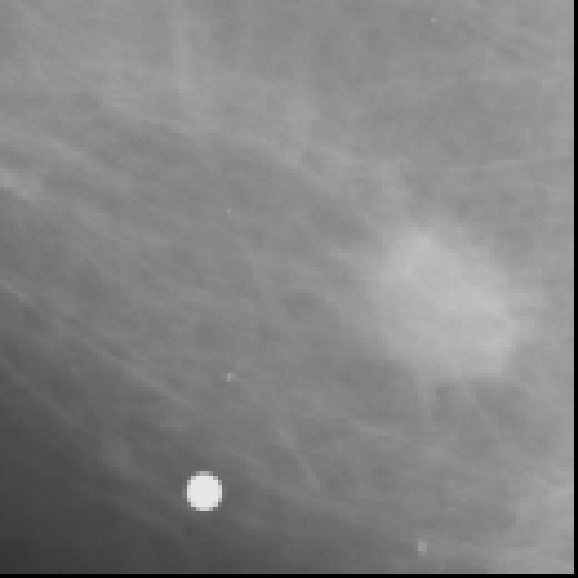}
	\end{subfigure}
\caption{Comparison of segmentation of analyzed  techniques. (a) Region of Interest; (b)Proposed Method; (c) Topographic; (d) Wavelet; (e) BEMD; (f) BMCS; (g) MCW.}
\label{fig:segAll}
\end{figure}

This section shows the results of the state of the art techniques applied to segment lesions of mammograms, from IRMA database, from fat transparent and fibroid tissues, corresponding to 685 images divided into circumscribed, spiculated and other mass. The proposed approach was compared with state of the art techniques and the results of the segmentation of each technique were evaluated through the metric described previously. Figure \ref{fig:segAll} shows the results of segmentation of all techniques analyzed  for the images of IRMA database.

Figure \ref{fig:segAll} shows 8 patches from IRMA database and the segmentation of each analyzed technique. The region of interest from the database is shown in column (a), where in the other columns the segmentation of each technique is represented in green. As can be observed, the proposed method and Topographic approach obtained a well-defined segmentation for most of cases of Figure \ref{fig:segAll}.

As described in the evaluation section, the segmented images of each technique were submitted to a classifier to identify the type of lesion according to its features of shape and margin. The classifier used was a MLP, which classifies the region of interest in benign or malignant. The analysis was separated according to the type of tissue and for each tissue it was divided analyzing two scenarios: a)circumscribed and spiculated lesions and b) circumscribed, spiculated and other masses. The results of classification for each scenario described are shown in Table \ref{tab_one}.

\begin{table}[htb]
\centering
\caption{Classification accuracy rate using the segmented images of the analyzed techniques, for fat and fibroid tissue.}
\label{tab_one}
\footnotesize
\begin{tabular}{lllll}
\hline
\multirow{2}{*}{Techniques} & \multicolumn{2}{c}{Fat Tissue}        & \multicolumn{2}{c}{Fibroid Tissue}       \\ \cline{2-5} 
                            & circ.+spic. & circ.+spic.+other & circ.+spic. & circ.+spic.+other \\ \hline
BEMD                        & 75.93$\pm$3.47\%         & 75.32$\pm$3.60\%           & 78.22$\pm$3.8\%         & 75.64$\pm$3.11\%          \\
BMCS                        & 76.15$\pm$3.21\%         & 72.07$\pm$2.41\%           & 85.50$\pm$4.42\%          & 72.37$\pm$2.96\%          \\
MCW                         & 69.52$\pm$3.49\%         & 70.01$\pm$3.18\%           & \textbf{86.17$\pm$3.47\%}         & 70.91$\pm$4.29\%          \\
Proprosed                   & \textbf{85.83$\pm$5.67\%}         & 75.93$\pm$3.94\%           & 84.30$\pm$1.95\%        & 72.48$\pm$3.83\%          \\
Topographic                 & 76.82$\pm$4.85\%         & \textbf{77.00$\pm$4.15\%}           & 84.61$\pm$5.94\%         & \textbf{76.81$\pm$4.61\%}      \\
Wavelet                     & 81.64$\pm$5.35\%         & 75.48$\pm$5.51\%           & 84.84$\pm$5.96\%         & 76.32$\pm$5.97\%         \\ \hline
\end{tabular}
\end{table}

\begin{figure}[!htb]
\centering
\includegraphics[width=0.85\textwidth]{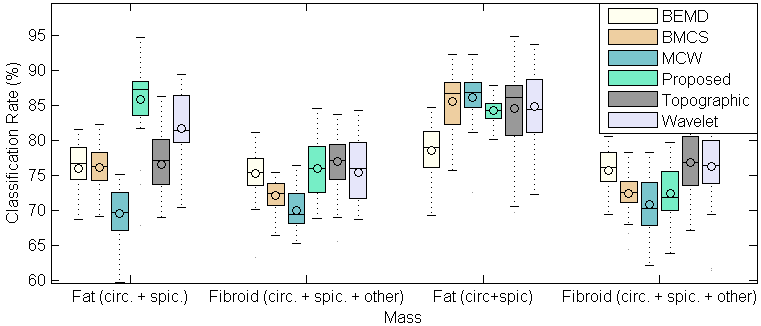}
\caption{Boxplot of the classification rate of the techniques analyzed.}
\label{fig:box1}
\end{figure}

The boxplot of results showed in Table \ref{tab_one} is illustrated in Figure \ref{fig:box1}.

As can be observed in Figure \ref{fig:box1}, the major difference of results between techniques was in the first scenario, when using fat tissue and only circumscribed and spiculated masses, where the proposed approach had higher classification rate. To evaluate if the results were statistically different, it was performed a hypothesis test. The hypothesis test was done using Student's t-test \citep{samuels2012statistics}, considering null hypothesis with equal population mean, using a confidence level of 95\%. The Student's t-test was done comparing the results of the proposed technique against the other ones. The results of this test is shown in Table \ref{tab_pvalue}.

\begin{table}[htb]
\centering
\caption{P-value of Student's t-test comparing the proposed approach with the analyzed techniques.}
\label{tab_pvalue}
\footnotesize
\begin{tabular}{lllll}
\hline
\multirow{2}{*}{Comparison} & \multicolumn{2}{c}{Fat Tissue}        & \multicolumn{2}{c}{Fibroid Tissue}       \\ \cline{2-5} 
                            & circ.+spic. & circ.+spic.+other & circ.+spic. & circ.+spic.+other \\ \hline
Topographic                 & \textbf{1.4418E-08}         & 0.3082          & 0.7908         & \textbf{0.0002}      \\
Wavelet                     & \textbf{0.00469}         & 0.7162          & \textbf{0.0067}         & \textbf{0.0046} \\
BEMD                        & \textbf{1.3128E-10}         & 0.5363           & \textbf{5.4451E-10}         & \textbf{0.0068}         \\
BMCS                        & \textbf{1.9458E-10}         & \textbf{3.4508E-05}           & 0.1830          & 0.9035         \\
MCW                         & \textbf{6.9151E-18}         & \textbf{3.5855E-08}          & 0.0131         & 0.1405          \\
 \hline
\end{tabular}
\end{table}

The bold values in Table \ref{tab_pvalue} represent the situations in which the null hypothesis were rejected. This means that the proposed approach and the compared technique are statistically different. For the other cases it means that they are statistically similar.  

Although most of the techniques obtained a good classification rate using the Zernike moments to the segmented images, it is not guaranteed that the edges of the segmented image are well defined. Therefore, another analysis was done focusing on the quality of segmentation and the classification rate using only the well segmented images. This does not invalidate the first analysis, because results showed that the obtained segmentation is suitable for the correct identification of tumors in benign or malignant, once it provides the contour features necessary to classify the tumor. However, for a more specific analysis about the quality of segmentation, it was separated the well segmented images obtained for each technique. For this purpose, it was considered that a well segmented image was that ones in which more than 50\% of the edges were not touching the edges of the region of interest. This decision was made because it was assumed that it is necessary to have more than 50\% of a well-defined edge to be considered as a good segmentation. Based on this, the next analysis evaluates the classification accuracy rate based only on the selected well-defined segmentations. This aims to analyze the amount of images selected as good segmentation for each technique and the classification rate for this selection. The process of selection was performed automatically based on edges of the tumor. The results for this analysis is showed in Table \ref{tab_final}.

\begin{table}[htb]
\centering
\caption{Classification rate of the analyzed techniques, when using the segmented images with a well-defined margin.}
\label{tab_final}
\resizebox{\textwidth}{!}{%
\begin{tabular}{lllllllll}
\hline
\multirow{3}{*}{Technique} & \multicolumn{4}{c}{Fat Tissue} & \multicolumn{4}{c}{Fibroid Tissue} \\ \cline{2-9} 
 & \multicolumn{2}{c}{circ. + spic.} & \multicolumn{2}{c}{circ. + spic. + other} & \multicolumn{2}{c}{circ. + spic.} & \multicolumn{2}{c}{circ. + spic. + other} \\ \cline{2-9} 
 & \multicolumn{1}{c}{\begin{tabular}[c]{@{}c@{}}Classification\\ Rate\end{tabular}} & Selection & \multicolumn{1}{c}{\begin{tabular}[c]{@{}c@{}}Classification\\ Rate\end{tabular}} & Selection & \multicolumn{1}{c}{\begin{tabular}[c]{@{}c@{}}Classification\\ Rate\end{tabular}} & Selection & \multicolumn{1}{c}{\begin{tabular}[c]{@{}c@{}}Classification\\ Rate\end{tabular}} & Selection \\ \hline
BEMD & - & 20/152 & - & 46/345 & - & 13/198 & - & 23/340 \\
BMCS & - & 12/152 & - & 28/345 & - & 10/198 & - & 33/340 \\
MCW & 85.69$\pm$6.03\% & 78/152 & 86.12$\pm$4.23\% & 169/345 & 89.77$\pm$4.41\% & 81/198 & 90.11$\pm$3.44\% & 128/340 \\
Proposed & 91.28$\pm$2.96\% & 87/152 & 88.34$\pm$5.03\% & 186/345 & 89.27$\pm$4.12\% & 125/198 & 85.52$\pm$4.39\% & 211/340 \\
Topographic & 84.20$\pm$5.33\% & 143/152 & 83.49$\pm$4.34\% & 317/345 & 86.97$\pm$5.21\% & 185/198 & 81.56$\pm$6.35\% & 313/340 \\
Wavelet & 89.81$\pm$3.29\% & 67/152 & 90.16$\pm$3.83\% & 125/345 & 89.76$\pm$3.95\% & 55/198 & 90.60$\pm$4.13\% & 87/340 \\ \hline
\end{tabular}
}
\end{table}

In Table \ref{tab_final}, the BEMD and BMCS approaches do not have a classification rate because the amount of selected images was too low to the classifier training process. That means that few images of BEMD and BMCS had more than 50\% of the edges well defined.

\section{Discussion}

This work presents a methodology for delineating masses on ROIs of digital mammograms, aiming to help the specialist in the identification of the lesion. The delineation approach is based on a modification of the seeded region based method GrowCut. In this modification the updated evolution rule employees a fuzzy Gaussian membership function. This modification reduces the effort of seeds selection, once only the foreground seeds are necessary to estimate the region of the lesion. Furthermore, it facilitates the use of unsupervised methods to select the seeds, as proposed in this work. In this section we discuss the results showed previously, analyzing the performance of the algorithms and the results obtained. 

In Figure \ref{fig:segAll}, the proposed technique, in column \emph{b}, obtained a contour close to the edges of the tumor for the images presented. The Topographic approach, in column \emph{c}, also obtained a good segmentation for most of cases, but it was not so well defined in some cases, like in the first, second and last image, were the segmentation was wrong. The Wavelet based approach, in column  \emph{d} did not obtain a good segmentation for some cases where the contour was ill-defined. The other techniques, in column \emph{e}, \emph{f} and \emph{g} did not segment well for most of the cases, segmenting the entire region of interest. The analysis was made for the 685 images, but the examples shown in Figure \ref{fig:segAll} illustrates the quality of segmentation of the proposed segmentation model.

Table \ref{tab_one} shows that the proposed approach achieves a higher accuracy when using the classifier to classify the segmented images in benign or malignant, applied to fat tissues and using circumscribed and spiculated masses. When including also other masses, in the third column, the average of classification is close to the Topographic and Wavelet approaches. For fibroid tissues, all techniques, except BEMD, obtained similar performance when considering only circumscribed and spiculated masses. With the addition of other masses for fibroid tissue, the Topographic approach obtained higher result. Can also be observed that not necessarily an algorithm that detect masses on fibroid tissue has a higher performance compared to one that identifies masses in fat tissue. For the cases analyzed, the MCW approach had a better performance for fibroid tissue when compared to the proposed approach, whereas for fat tissue the proposed approach had better results.

In Table \ref{tab_pvalue}, the results of second column means that for fat tissue, using only circumscribed and spiculated masses, the proposed approach has statistically different results when compared to the other techniques. Therefore, can be said that the proposed technique obtained a high accuracy when used its segmented images with the classifier. For fat tissue, including other masses, it was statistically different when compared to BMCS and MCW approaches. However, it had no statistical evidence that it was different from Topographic, Wavelet and BEMD approaches. This means that despite Topographic approach had a higher average classification rate, it was not statistically different from the proposed method. For fibroid tissue, the Wavelet based technique was statistically different, having higher slight higher accuracy when using circumscribed and spiculated masses. When adding other masses, Topographic, Wavelet and BEMD approach were statistically superior.

Table \ref{tab_final} shows the classification rate and the amount of selected images, for each technique, for fat and fibroid tissues. The dataset for each technique is the same, however it was performed a selection of images based on the quality of segmentation. This is done because the objective is to evaluate the confidence level of each technique, showing a relation between the amount of well segmented images and its accuracy for this set. If a technique has a high accuracy when using the classifier, this indicates that the quality of segmentation is high. Therefore, if a technique has few well segmented images, but the classification rate is high, this means a high quality of segmentation and the confidence level of its segmentation is high. On the other hand, if a technique has several well segmented images, but the classification rate is low, the confidence level of its segmentation is low. In the second column of Table \ref{tab_final}, the proposed approach reaches 91.28\% of classification rate, having 87 of 152 images considered as having a well-defined contour, as showed in the third column of Table \ref{tab_final}. The Topographic approach, despite having a higher number of images selected, it had a lower classification rate. Therefore, the analysis shows the relation between the amount of well segmented images and the classification rate when used only the images considered with well defined contour.  From the first case of fat tissue, the proposed approach had lower number of well segmented images, but the classification rate shows that the quality of segmentation was better. The second case, where the circumscribed, spiculated and other masses were analyzed, for fat tissue, the Wavelet based approach had a higher classification rate. However, the proposed approach had a close rate with a higher number of well segmented images. For fibroid tissue the proposed and the Topographic approach had a good tradeoff between classification rate and number of selected images.

Experimental outcomes indicate that using the segmentation generated by the proposed method will lead to a better classification rate for fat tissues. This represents that the segmentation provided better characteristics to the classifier distinguish between tumor and not tumor for this type of tissue. This results also suggest that the quality of segmentation was better when using Fuzzy GrowCut.  One of the aspects that makes the Fuzzy GrowCut obtain better segmentation results is that the method is less dependent on a correct initialization when compared to state-of-art techniques. Therefore, even if the unsupervised step of generation of seed is not perfect, the algorithm can provide an accurate segmentation. Moreover, it does not rely on a threshold value, as found in Topographic Approach and BMCS. The reduction of dependence on initialization has high implications on segmentation tasks where user knowledge is required, but not guaranteed that is correct. Furthermore, the proposed method can be extended to other kind of medical images. This explanation was added to the discussion section.

	One of the main strengths of the proposed method is that it is flexible to the seeds' initialization. This happens because the propagation of seeds is based on the center of mass of all seeds, and not on the seeds individually. With the addition of a Fuzzy membership function, the segmentation process becomes more flexible, different from state-of-art techniques which uses the seeds as reliable information. Therefore, besides reducing specialist knowledge necessary to initialization and removing the need of selecting background seeds, it has as consequence the reduction of weight related to the correct generation of seeds in an unsupervised approach. In state of the art seed based techniques, such as Random Walks, it is hard to adapt the method to an unsupervised approach, once the automatic generation of seeds cannot contain incorrect labelling.
	
	On the other hand,  of the weakness of the proposed approach is that it requires more computational time compared to state of the art techniques.

\section{Conclusion} \label{sec:conclusion}

Herein this work we proposed a new approach to segment masses in digital mammography images. This approach is based on a semi-supervised modification of GrowCut segmentation algorithm, using fuzzy Gaussian membership functions in the new evolution rule. With such a fuzzy function we were able to deal with complex non-defined tumor boundaries, as our qualitative results demonstrate. In order to surpass GrowCut limitation of needing human intervention at selecting internal and external points to train the segmentation method, we included a non-supervised previous stage with the ability to automatically select internal points using the classical simulated annealing algorithm. Our fuzzy approach avoids the need of selecting external points.

The proposed technique was evaluated with 685 images from the IRMA database and compared with the following techniques:  BEMD, BMCS, LBI, MCW, Topographic Approach and Wavelet Analysis. The evaluation was done applying the Zernike moments on the segmented images and using the MLP to classify the images in benign or malignant. This estimates the quality of segmentation, since the database does not provided the ground truth. The evaluation was performed for images of fat tissue and fibroid tissue, using circumscribed, spiculated and other masses.

Results showed that the proposed approach had better results on average for fat tissue, obtaining 85.83\% of classification rate. We also employed Student's t-test to identify differences among the several methods we implement, and results pointed that our approach is significantly different from others in this scenario. When including other masses, for fat tissue, the method we proposed can be considered statistically equivalent to others. For fibroid tissue, the Wavelet and Topographic approaches had a slightly higher classification rate. When analyzing the quality of segmented images, the proposed approach obtained 91.28\% of classification rate for fat tissue, having a good tradeoff between well segmented images and classification rate. For fibroid tissues, the proposed approach had a good balance between classification rate and well segmented images, equivalent to the Topographic approach.

From these results, we can conclude that our semi-supervised modification of GrowCut, with automatic seed selection using simulated annealing and altered evolution rule based on fuzzy Gaussian membership functions, is feasible and suitable for breast tumor segmentation, mainly because it does not require additional human intervention once suspicious lesion areas are already clinically determined as input data in this application. Considering qualitative results, our proposal was able to perform good lesion segmentation for circumscribed and spiculated mammary lesions, having better qualitative segmentation than the state-of-the-art techniques we implemented, considering fat mammary tissues.

This approach can be extended for other biomedical image applications where fuzzy-boundaries objects have to be segmented. Regarding the computational effort, segmentation times could be minimized by using parallel architectures and strategies, due to the very parallel nature of the algorithms we proposed and used in this hybrid method, for example, classical simulated annealing and modified GrowCut.



\section{Acknowledgments}
The authors thank to the courtesy of Prof. Thomas Deserno, from the Department of Medical Informatics, RWTH Aachen University, Germany, which kindly provided access to the IRMA dataset. 



\bibliographystyle{elsarticle-harv}
\bibliography{referencias}







\end{document}